\documentclass[manuscript,screen,nonacm]{acmart}

\AtBeginDocument{%
  }

\setcopyright{none}
\usepackage{booktabs}
\usepackage{pifont}
\usepackage[dvipsnames]{xcolor}
\usepackage{graphicx}
\usepackage[htt]{hyphenat}
\usepackage{array}
\usepackage{listings}
\usepackage{wrapfig}
\usepackage{caption}

\newcommand{\cmark}{\textcolor{green!60!black}{\boldmath$\checkmark$}}
\newcommand{\xmark}{\textcolor{BrickRed}{\boldmath$\times$}}

\usepackage{tikz}

\definecolor{myyellow}{RGB}{190,140,0}

\newcommand*\circledblue[1]{\tikz[baseline=(char.base)]{
            \node[shape=circle,draw=NavyBlue!100,fill=NavyBlue!10,thick,inner sep=1pt] (char) {\scriptsize\textsf#1};}}

\newcommand*\circledgreen[1]{\tikz[baseline=(char.base)]{
            \node[shape=circle,draw=ForestGreen!60,fill=ForestGreen!10,thick,inner sep=1pt] (char) {\scriptsize\textsf#1};}}
\newcommand*\circledred[1]{\tikz[baseline=(char.base)]{
            \node[shape=circle,draw=BrickRed!60,fill=BrickRed!10,thick,inner sep=1pt] (char) {\scriptsize\textsf#1};}}

\newcommand{\rqtag}[2]{%
  \tikz[baseline=(rq.base)]{
    \node[
      rounded corners=2.5pt,
      draw=#2,
      fill=#2!10,
      text=#2,
      thick,
      inner xsep=5pt,
      inner ysep=2.5pt
    ] (rq) {\small\bfseries\textsf{RQ#1}};
  }%
}

\newcommand{\researchquestion}[3]{%
  \par\medskip
  \noindent
  \rqtag{#1}{#2}\textbf{#3}
  \par\smallskip
}

\newcommand{\inlineRQ}[2]{%
  \mbox{\rqtag{#1}{#2}}\nobreak%
}

\newcommand{\dimension}[1]{\textsc{#1}}

\begin{document}

\title{\textsc{XAI-Arena}: Can LLMs Assess the Quality of XAI Explanations?}
%\\ A Reproducible LLM-Based Evaluation}

% \TODO{please add you, but also Alona, Najda, myself as authors/supervisors here with correct affiliation }

\author{Yanfei Hu Fleischhauer}
\affiliation{%
  \institution{LMU Munich}
  \city{Munich}
  \country{Germany}
}
\email{fayfayhu@gmail.com}

\author{Alona Zharova}
\affiliation{%
  \institution{Chair of Information Systems, School of Business and Economics, Humboldt-Universität zu Berlin}
  \city{Berlin}
  \country{Germany}
}
\email{alona.zharova@hu-berlin.de}

\author{Nadja Klein}
\affiliation{%
  \institution{Scientific Computing Center, Karlsruhe Institute of Technology}
  \city{Karlsruhe}
  \country{Germany}
}
\email{nadja.klein@kit.edu}

\author{Stefan Feuerriegel}
\affiliation{%
  \institution{MCML, LMU Munich}
  \city{Munich}
  \country{Germany}
}
\email{feuerriegel@lmu.de}


\begin{abstract}
Evaluating the quality of explanations produced by explainable AI (XAI) methods remains challenging because existing approaches often rely on subjective human judgment, limiting reproducibility, scalability, and comparability between studies. We examine whether LLMs can serve as a reproducible and scalable mechanism to make comparative assessments of the quality of XAI explanations. We introduce XAI-Arena, an LLM-as-a-judge framework for scalable, reproducible, multidimensional, and stakeholder-sensitive evaluation of XAI explanation quality. XAI-Arena then allows us to compare XAI explanations along various dimensions, namely, perceived simplicity, clarity, task adequacy, trust calibration, actionability, transparency, faithfulness, and overall interpretability. We then benchmark XAI explanation methods across various datasets, machine learning models, and stakeholder personas. Human validation shows a strong positive association between LLM-generated and human ratings (Spearman's $\rho=.693$, $p<.001$). Together, LLM-based evaluations can capture systematic differences in XAI explanation quality and provide a scalable and reproducible framework for comparative assessment of XAI explanations.
\end{abstract}


\begin{CCSXML}
<ccs2012>
   <concept>
       <concept_id>10003120</concept_id>
       <concept_desc>Human-centered computing</concept_desc>
       <concept_significance>500</concept_significance>
       </concept>
   <concept>
       <concept_id>10003120.10003121.10011748</concept_id>
       <concept_desc>Human-centered computing~Empirical studies in HCI</concept_desc>
       <concept_significance>500</concept_significance>
       </concept>
   <concept>
       <concept_id>10010147.10010178</concept_id>
       <concept_desc>Computing methodologies~Artificial intelligence</concept_desc>
       <concept_significance>500</concept_significance>
       </concept>
   <concept>
       <concept_id>10010147.10010257</concept_id>
       <concept_desc>Computing methodologies~Machine learning</concept_desc>
       <concept_significance>300</concept_significance>
       </concept>
 </ccs2012>
\end{CCSXML}

\ccsdesc[500]{Human-centered computing}
\ccsdesc[500]{Human-centered computing~Empirical studies in HCI}
\ccsdesc[500]{Computing methodologies~Artificial intelligence}
\ccsdesc[300]{Computing methodologies~Machine learning}

\keywords{Explainable AI, explanation quality, interpretability, user study, LLM-as-a-judge, human-centered evaluation, stakeholder perspectives, reproducibility}

% \received{20 February 2026}
% \received[revised]{12 March 2026}
% \received[accepted]{5 June 2026}

\maketitle

\section{Introduction}

\begin{figure}
    \centering
    \includegraphics[width=\textwidth,height=0.60\textheight,keepaspectratio]{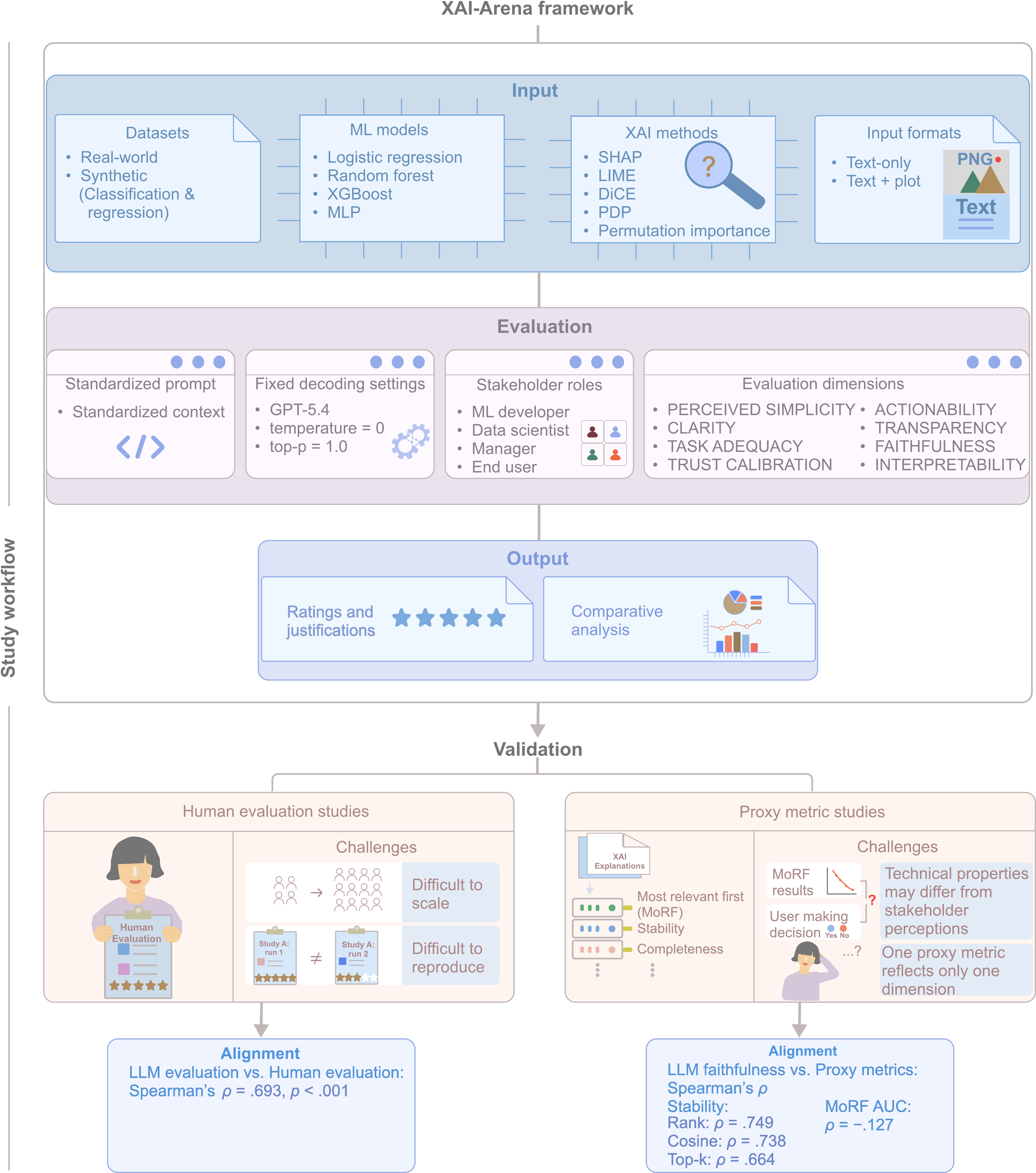}
    \caption{Our proposed \textsc{XAI-Arena} framework. It addresses limitations of existing XAI evaluation approaches by proposing an LLM-as-a-judge framework for scalable, multidimensional, and stakeholder-sensitive assessment of XAI explanation quality.}
    \Description{A visual overview of XAI-Arena. The framework receives explanations generated across real-world and synthetic datasets, classification and regression tasks, multiple machine learning models, five XAI methods, and text-only or text-plus-plot formats. A standardized prompt and fixed decoding settings are used to evaluate each explanation from four stakeholder perspectives across eight quality dimensions, producing ratings and justifications for comparative analysis. The framework addresses limitations of human evaluation, which is difficult to scale and reproduce, and technical proxy metrics, which capture only selected properties and may differ from human perceptions. External validation shows strong alignment between LLM and human ratings, with Spearman's rho equal to 0.693, and strong alignment between LLM-rated faithfulness and the stability metrics rank stability, cosine similarity, and top-k overlap.}
    \label{fig:visual_abstract}
\end{figure}

Understanding how machine learning models arrive at their predictions is essential in many practical applications. The underlying reasons may be legal, organizational, or technical, such as ensuring regulatory compliance, auditing model decisions, understanding and improving model behavior, or building trust among managers and end users. One way to achieve this is through explainable AI (XAI), which provides explanations for model predictions \cite{doshi2017towards, adadi2018peeking, guidotti2018survey}. XAI helps users examine model behavior \cite{samek2017explainable}, identify and debug model issues \cite{balayn2022can}, and communicate why a model produced a specific output \cite{miller2019explanation}.

However, evaluating the quality of XAI explanations remains challenging \cite{martens2025bewareexplanationsai}. Existing evaluations often rely on human judgment, which can provide rich insights but is costly to collect and difficult to scale \cite{miller2019explanation, liao2021human,bhatt2020explainable}. Moreover, human judgment can vary due to differences in tasks or expertise of participants \cite{buccinca2020proxy, kim2024human, mohseni2021multidisciplinary}. Automated proxy metrics provide a more scalable alternative, but  typically measure narrow technical properties, such as fidelity \cite{doshi2017towards}, stability \cite{alvarez2018robustness}, and completeness \cite{vilone2021notions, nauta2023anecdotal}, rather than broader dimensions of explanation quality. In particular, it is unclear whether such proxy metrics align with how stakeholders perceive or use XAI explanations \cite{nauta2023anecdotal, vilone2021notions}. As a result, comparative assessments of XAI explanation quality often remain difficult to conduct in a systematic and reproducible manner.

Comparative assessment of XAI explanations is further challenged by several degrees of freedom in the evaluation setting. First, XAI methods differ in scope. Some methods explain individual predictions, whereas others summarize model behavior globally. For example, SHAP and LIME provide local feature attribution explanations for individual predictions \cite{lundberg2017unified,ribeiro2016should}, whereas permutation importance summarizes feature relevance at the model level. Second, explanations differ in format. Some methods provide textual and tabular outputs, while others offer visual representations (e.g., partial dependence plots). Third, explanations are interpreted by different stakeholders \cite{doshi2017towards, miller2019explanation, liao2021human, senoner2024explainable, brennen2020people}, such as developers, data scientists, managers, and end users, who differ in expertise and decision context and thus may perceive explanation quality differently. Figure~\ref{fig:visual_abstract} provides an overview of the proposed framework and its validation against human evaluations and technical proxy metrics.

In this paper, we analyze whether large language models (LLMs) can serve as reproducible and scalable evaluators of XAI explanation quality. For this purpose, we introduce \textsc{XAI-Arena}, a controlled LLM-as-a-judge framework in which a single LLM, under fixed prompting and decoding settings, is used to assess the quality of XAI explanations within a unified evaluation protocol. We evaluate five widely used XAI methods: SHAP \cite{lundberg2017unified}, LIME \cite{ribeiro2016should}, DiCE \cite{mothilal2020explaining}, partial dependence plots (PDP) \cite{goldstein2015peeking, friedman2001greedy}, and permutation importance \cite{fisher2019all}. The explanations are assessed from the perspective of four stakeholder personas: ML developer, data scientist, manager, and end user. We compare explanations along eight established dimensions of explanation quality from prior work on human-centered XAI \cite{doshi2017towards, hoffman2018metrics, miller2019explanation, liao2021human, bhatt2020explainable, mohseni2021multidisciplinary, buccinca2020proxy, nauta2023anecdotal}: \dimension{perceived simplicity}, \dimension{clarity}, \dimension{task adequacy}, \dimension{trust calibration}, \dimension{actionability}, \dimension{transparency}, \dimension{faithfulness}, and overall \dimension{interpretability}. As a result, \textsc{XAI-Arena} provides an evaluation framework to systematically compare the quality of XAI explanations in a scalable and reproducible manner. Importantly, we do not treat LLM-based ratings as a direct substitute for human assessments; rather, using human ratings, we examine whether they recover systematic differences and preference patterns that can guide choices among XAI explanations.

\begin{figure}
  \centering
  \includegraphics[
      width=0.90\textwidth,
      % height=0.52\textheight,
      keepaspectratio
  ]{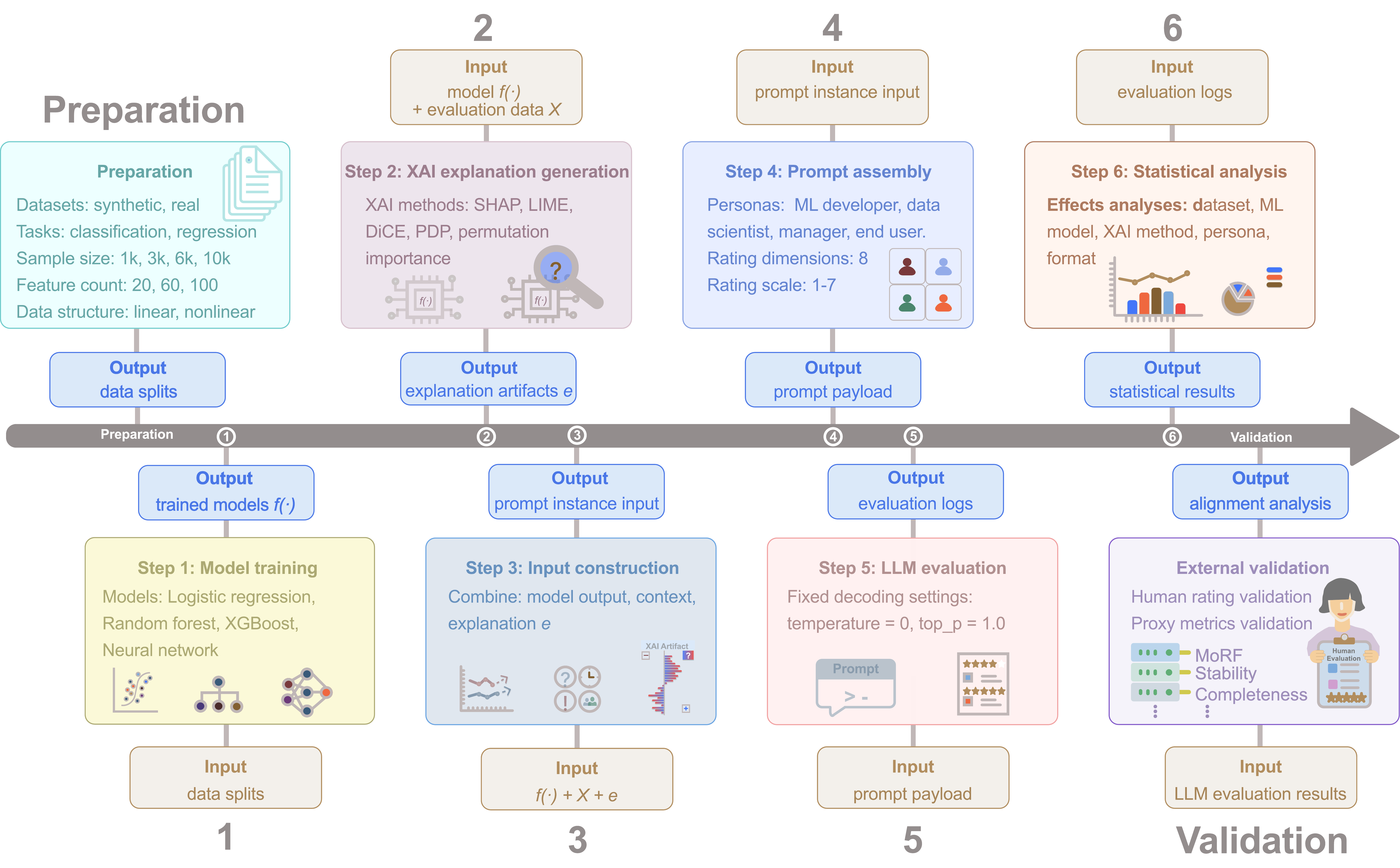}
  \caption{Overview of the \textsc{XAI-Arena} framework, showing the six-step evaluation pipeline from model training and explanation generation to prompt construction, LLM evaluation, and downstream analysis, together with external validation through human evaluation and proxy metrics.}
  \Description{Flow diagram of the six-step \textsc{XAI-Arena} framework, from model training and XAI explanation generation through prompt construction and LLM-based evaluation to downstream analysis, with external validation using human evaluation and proxy metrics.}
  \label{fig:xai_arena_pipeline}
\end{figure}

\smallskip
\noindent\textbf{Contributions:}
This paper makes the following contributions:\footnote{Code and data is available: \url{https://anonymous.4open.science/r/xai-arena/}}
\begin{description}
\item[\circledblue{1}] We introduce \textsc{XAI-Arena}, the first LLM-as-a-judge framework for assessing the multi-dimensional quality of XAI explanations. Our framework provides a scalable and reproducible approach for benchmarking the explanation quality in future XAI research.
\item[\circledblue{2}] We conduct a large-scale comparative study across different machine learning models, XAI explanation methods, datasets, stakeholder roles, and explanation formats.
\item[\circledblue{3}] We provide empirical evidence on how XAI explanation quality varies across explanation methods, stakeholder roles, and explanation formats. Thereby, we identify systematic preferences for XAI explanations across application patterns. 
\end{description}

\section{Related Work}

Prior work in XAI has developed numerous methods to make model behavior more transparent and thereby support an accountable, trustworthy use of AI systems \cite{doshi2017towards, adadi2018peeking, guidotti2018survey}. Below, we provide a short overview of existing (1)~key XAI methods, (2)~evaluation strategies to assess explanation quality, and (3)~recent applications following the LLM-as-a-judge paradigm. 

\subsection{Explainable AI Methods}

% defintion of XAI
% \noindent\textbf{Explanation techniques.} 
We survey key XAI methods\footnote{We use the term \textit{explanation} broadly to refer to any output produced by an XAI method. We recognize that the terminology around explainability and interpretability varies across the literature \cite{lipton2018mythos, doshi2017towards}, and we here adopt a pragmatic use to broadly refer to the resulting explanation artifact.} while we refer readers to \cite{guidotti2018survey, adadi2018peeking, molnar2020interpretable} for more comprehensive overviews. Common methods in practice include: (i)~\textbf{SHAP} \cite{lundberg2017unified}, which uses Shapley values from cooperative game theory to attribute feature importance consistently across models; (ii)~\textbf{LIME} \cite{ribeiro2016should}, which fits local surrogate models around individual predictions to approximate model behavior; (iii)~counterfactual explanations such as \textbf{DiCE} \cite{mothilal2020explaining}, which identify minimal input changes needed to alter the model’s output; and (iv)~global visualization techniques such as \textbf{PDP} \cite{friedman2001greedy, goldstein2015peeking} and \textbf{permutation importance} \cite{fisher2019all, breiman2001random}. PDP illustrates how model predictions change with respect to individual features, while permutation importance measures the contribution of each feature to the overall predictive accuracy \cite{breiman2001random, strobl2008conditional, molnar2020interpretable, hooker2021unrestricted}. 

In this paper, we focus on five widely used methods (SHAP, LIME, DiCE, PDP, and permutation importance), which represent complementary explanation paradigms used in practice. One difference is that some explain individual predictions locally, while others summarize overall model behavior globally. They also vary in representation (textual, visual, or hybrid) and abstraction level, ranging from feature-level explanations to higher-level summaries of model behavior.

% Literature notes: Related Work - XAI methods 

\subsection{Evaluation of XAI Explanations}
\label{sec:xai_evaluation}

Despite extensive research on XAI methods, systematic comparative evaluations of explanation quality remain limited \cite{gilpin2018explaining, arya2019one, covert2021explaining}. Existing studies often focus on specific methods, tasks, or evaluation settings, which makes it difficult to assess how explanation quality varies across machine learning models, datasets, explanation methods, output formats, and stakeholders. For instance, an attribution method that yields clear and stable explanations for a linear model may produce noisier or less interpretable explanations in a highly nonlinear setting \cite{slack2020fooling}. This motivates the development of \textsc{XAI-Arena} as a comparative evaluation framework in this paper.

% \noindent\textbf{Evaluation Dimensions.} 

Several dimensions can be used to assess the quality of XAI explanations. Early work emphasized two main outcomes: faithfulness, which captures how accurately an explanation reflects the model's reasoning, and interpretability, which captures how understandable it is for human users \cite{doshi2017towards, gilpin2018explaining}. Subsequent research expanded these perspectives into broader, multidimensional frameworks that combine technical and human-centered criteria, including clarity, completeness, plausibility, and usefulness \cite{hoffman2018metrics, miller2019explanation, liao2021human, chromik2021think, kim2023help}. Building on this line of work, we adopt eight complementary evaluation dimensions: \circledred{1}~\dimension{perceived simplicity}, \circledred{2}~\dimension{clarity}, \circledred{3}~\dimension{task adequacy}, \circledred{4}~\dimension{trust calibration}, \circledred{5}~\dimension{actionability}, \circledred{6}~\dimension{transparency}, \circledred{7}~\dimension{faithfulness}, and \circledred{8}~\dimension{interpretability}. Note that we use \dimension{perceived simplicity} rather than perceived complexity so that higher ratings consistently indicate better explanation quality. Together, these dimensions should enable a comprehensive assessment of XAI explanation quality.

% Literature notes: Related Work - Evaluation Dimensions 

Existing human-centered evaluations of XAI explanations vary widely in design and methodology. A common choice is to rely on user studies in which participants rate explanation quality on Likert scales, perform decision support tasks, or provide qualitative feedback through interviews and think-aloud protocols \cite{doshi2017towards, mohseni2021multidisciplinary, liao2021human, buccinca2020proxy}. Participants typically include both domain experts and non-experts, whose interpretations depend on prior technical experience and task framing \cite{liao2021human, miller2019explanation, ehsan2024xai}. However, broad role- or expertise-based categories may obscure differences in stakeholder knowledge and explanation needs~\cite{suresh2021beyond}. While such studies provide fine-grained insights into how users interpret XAI explanations and calibrate trust in model predictions, they are expensive, time-consuming, and difficult to reproduce at scale.  

% Literature notes: Related Work - Human-Centered Evaluation

Proxy metrics provide automated, quantitative measures for evaluating XAI explanations without human input \cite{doshi2017towards, nauta2023anecdotal}. Fidelity-based measures assess how closely an explanation aligns with model behavior, while criteria such as completeness, stability, sensitivity, MoRF (Most Relevant First), and compactness capture different technical aspects of explanation quality \cite{alvarez2018robustness, nauta2023anecdotal}. For example, MoRF evaluates how strongly the model output changes when the most important input features are removed \cite{samek2016evaluating}, and stability measures how XAI explanations change under small variations in the input \cite{alvarez2018robustness}. However, proxy metrics remain limited because they primarily benchmark technical properties of explanations rather than how people interpret, compare, or use them. In particular, they do not capture human responses to presentation format, visual explanations, stakeholder context, or perceived usefulness. As a result, proxy metrics are \underline{not} suitable to assess the XAI explanation quality in a human-centered way, which is our focus \cite{sokol2024does}.

% Literature notes: Related Work - Proxy Metrics

\subsection{LLMs as Evaluators}

LLM-based evaluations, often referred to as ``LLM-as-a-judge'', have been applied across various tasks, including text summarization, recommendation explanations, and educational feedback \cite{kocmi2023large, belouadah2025evaluating, zhang2024large, ma2026synthetic, ma2026skillgen}. Empirical studies report moderate to strong correlations between LLM-based and human judgments, suggesting that LLMs can approximate human evaluation in many subjective assessment tasks \cite{wang2025can,seo2025large,zhang2024large}. At the same time, evaluation performance depends on factors such as prompt phrasing and the underlying LLM model, which highlights the importance of standardized and transparent prompting procedures \cite{bavaresco2025llms,li2025generation}. More broadly, the emerging LLM-as-a-judge paradigm investigates whether LLMs can approximate human judgments or serve as scalable evaluators for complex outputs \cite{chiang2023can,bavaresco2025llms,gilardi2023chatgpt}. However, this literature has not focused on the evaluation of XAI explanation quality.

A related but distinct line of work uses LLMs to generate or verbalize model explanations in natural language, for example, through interactive explanation systems (i.e., ``talk to the model''  \cite{liao2020questioning}). In contrast, our work uses LLMs to evaluate explanations generated by external XAI methods.

\smallskip
\noindent\textbf{Research gap:} Prior work typically uses human raters or proxy metrics to assess XAI explanation quality (Table~\ref{tab:streams}), but scalable, reproducible frameworks for human-centered assessments are missing. To the best of our knowledge, \textsc{XAI-Arena} is the first LLM-as-a-judge framework designed for this purpose.

\begin{table}
\centering
\footnotesize
\caption{Comparison of approaches for assessing the quality of XAI explanations.
\cmark = supported, \xmark = not supported.}
\label{tab:streams}

\renewcommand{\arraystretch}{1.15}
\setlength{\tabcolsep}{3pt}

\begin{tabular}{p{5.8cm}ccccc}
\toprule

\textbf{Approach} &
\textbf{Human-} &
\textbf{Scalable} &
\textbf{Repro-} &
\textbf{Multi-} &
\textbf{Stakeholder} \\

&
\textbf{centered} & &
\textbf{ducible} &
\textbf{dimension} &
\textbf{roles} \\

\midrule

Human-centered evaluation
\cite{doshi2017towards, liao2021human, mohseni2021multidisciplinary}
& \cmark & \xmark & \xmark & \cmark & \cmark \\

Proxy metrics
\cite{alvarez2018robustness, nauta2023anecdotal}
 & \xmark & \cmark & \cmark & \xmark & \xmark \\

\midrule
\textbf{\textsc{XAI-Arena} (\emph{this work})}
& \cmark & \cmark & \cmark & \cmark & \cmark  \\

\bottomrule
\end{tabular}

\end{table}

% Literature notes: Related Work - Research gap

%%%%%%%%%%%%%%%%%%%%%%%%%%%%%%%%%%%%%%%%%%%%%%%%%%%%%%%

\section{The \textsc{XAI-Arena} Framework}

This section presents \textsc{XAI-Arena}, a reproducible LLM-as-a-judge framework for assessing the quality of XAI explanations. \textsc{XAI-Arena} follows six steps: \circledgreen{1}~model training, \circledgreen{2}~XAI explanation generation, \circledgreen{3}~input construction, \circledgreen{4}~prompt assembly, \circledgreen{5}~LLM evaluation, and \circledgreen{6}~statistical analysis (see Figure~\ref{fig:xai_arena_pipeline}). The \textsc{XAI-Arena} framework is general and can be instantiated with different XAI methods, machine learning models, datasets, stakeholder personas, and evaluation dimensions.

\subsection{Problem description}

\textsc{XAI-Arena} defines an evaluation protocol that is general and not specific to any dataset, model, or XAI method. In the following, we describe each step and specify how inputs, prompts, and outputs are constructed to enable controlled and reproducible assessment of XAI explanations across stakeholder roles and explanation formats.

\textbf{Input.}
Let $f$ denote a trained model, and let $X$ denote the input data available for constructing an explanation artifact. For local explanations, let $x \in X$ denote the input instance being explained, and let $f(x)$ denote the corresponding model prediction. We denote the XAI explanation method by $e$ (e.g., SHAP, LIME, etc.). Depending on the XAI method, $e$ may be a local explanation $e(x,f)$ for a specific instance, or a global explanation $e(X,f)$ summarizing model behavior over a dataset or sample. We call $e(x,f)$ and $e(X,f)$, respectively, the explanation artifact, and simply write $e'$ as short-hand notation. The evaluation input additionally specifies a stakeholder persona $p \in \mathcal{P}$, which defines the role from whose perspective the explanation artifact is assessed (e.g., a line manager, an end user).

\textbf{Task.}
The task is to assess the quality of the explanation artifact $e'$ for the model behavior or prediction it explains, from the perspective of persona $p$. We define a set of evaluation dimensions $\mathcal{D} = \{d_1,\ldots,d_{n_d}\}$, where each dimension corresponds to one aspect of explanation quality derived from prior work on XAI evaluation; the concrete dimensions and their definitions are specified in the experimental setup (see Table~\ref{tab:dimensions}). Each dimension is rated on a seven-point Likert scale from 1 (very low) to 7 (very high), so that $r_i \in \{1,\ldots,7\}$ for each $d_i \in \mathcal{D}$.  

\textbf{Output.}
We denote the evaluator as a function $\textrm{LLM}_\textrm{judge}(\cdot)$ that maps the structured input to ratings and justifications:
\begin{equation}
\textrm{LLM}_\textrm{judge}(x, X, f, e', p) \rightarrow (\mathbf{r}, \mathbf{j}),
\end{equation}
where $\mathbf{r} = (r_1,\ldots,r_{n_d})^\top$ is the vector of ratings and $\mathbf{j} = (j_1,\ldots,j_{n_d})^\top$ is the corresponding vector of textual justifications for $n_d$ dimensions. Each pair $(r_i,j_i)$ corresponds to one evaluation dimension $d_i \in \mathcal{D}$  (e.g., \dimension{perceived simplicity}, \dimension{clarity}, \dimension{task adequacy}, etc.; defined later). For global explanations, $x$ is not used, and the evaluation is based on the model-level or dataset-level artifact $e(X,f)$. Because the persona $p$ is part of the evaluation input, the same explanation artifact $e'$ may receive different ratings across stakeholder roles, thus reflecting role-dependent interpretations of explanation quality. A further challenge is that many XAI artifacts are visual, which requires a \emph{multimodal} evaluation pipeline.

\begin{table}[h]
\centering
\caption{Illustrative input--output example for \textsc{XAI-Arena} using a single local explanation instance.}
\label{tab:io_example}
\small
\setlength{\tabcolsep}{4pt}
\renewcommand{\arraystretch}{1.12}
\begin{tabular}{
>{\raggedright\arraybackslash}p{2.8cm}
>{\raggedright\arraybackslash}p{4.9cm}
>{\raggedright\arraybackslash}p{5.8cm}
}
% \footnotesize
% \setlength{\tabcolsep}{3pt}
% \renewcommand{\arraystretch}{1.08}
% \begin{tabular}{
% p{2.0cm}
% p{2.8cm}
% >{\raggedright\arraybackslash}p{3.2cm}
% }
\toprule
\textbf{Step} & \textbf{Input} & \textbf{Output} \\
\midrule

\protect\circledgreen{1} Model training &
WDBC training split &
Trained model $f$ \\[2pt]

\protect\circledgreen{2} XAI explanation gen. &
Model $f$ and instance $x \in X_{\mathrm{test}}$ &
Prediction $f(x)=\texttt{benign}$; local SHAP artifact $e'$ \\[2pt]

\protect\circledgreen{3} Input construction &
Prediction $f(x)$ and artifact $e'$ &
Context: \texttt{"The tumor is predicted benign."} SHAP top-2 features: \texttt{mean\_radius}, \texttt{mean\_texture} \\[2pt]

\protect\circledgreen{4} Prompt assembly &
Context with persona $p$ and dimensions $\mathcal{D}$ &
Prompt excerpt: \texttt{"You are an end user Evaluate using a 1--7 scale."} \\[2pt]

\protect\circledgreen{5} LLM evaluation &
Standardized prompt &
Ratings $\mathbf{r}=(4,5,4,3,5,4,3,4)^\top$; justifications $\mathbf{j}$ \\[2pt]

\protect\circledgreen{6} Statistical analysis &
Ratings, justifications, metadata &
Logged record with model, XAI method, persona, explanation ID, ratings, and metadata \\

\bottomrule
\end{tabular}
\end{table}

\subsection{Workflow}
\label{sec:xai_arena_framework}

We now describe the workflow of XAI-Arena, which consists of six steps. Table~\ref{tab:io_example} illustrates these steps using a single local explanation instance.

\vspace{0.2cm}
\noindent\textbf{\protect\circledgreen{1} Model training.}
For each dataset and model combination, we train a machine learning model $f$ on the corresponding training split of the dataset. The trained model $f$ is stored to support traceable and reproducible explanation generation in subsequent steps.

\vspace{0.2cm}
\noindent\textbf{\protect\circledgreen{2} XAI explanation generation.} For each trained model $f$ and XAI explanation method $e$, we generate an explanation artifact for either a selected test instance or the relevant test data. For local XAI methods, we randomly sample a subset of held-out test instances $x \in X_{\mathrm{test}}$, compute the corresponding model predictions $f(x)$, and generate local explanation artifacts $e(x,f)$. For global XAI methods, we generate global explanation artifacts $e(X_{\mathrm{test}},f)$ that summarize the behavior of the trained model over the test data.

\vspace{0.2cm}
\noindent\textbf{\protect\circledgreen{3} Input construction.} For each explanation artifact generated by an XAI method, we construct the input to the LLM evaluator. The input combines the artifact $e'$ with textual context $C$ describing (i)~the prediction task, (ii)~the dataset, (iii)~the model type, (iv)~the XAI method used to generate the artifact, and, for local XAI methods, (v)~the model prediction $f(x)$ and (vi)~selected feature values of the instance $x$. The substantive content of the artifact $e'$ is preserved, while the representation is formatted consistently for the corresponding XAI method.

\vspace{0.2cm}
\noindent\textbf{\protect\circledgreen{4} Prompt assembly.}
The input to the LLM evaluator is inserted into a standardized prompt template (see Appendix~\ref{app:prompt}). Following best practice in prompt design \cite{lin2024write, giray2023prompt, feuerriegel2025using}, the template uses a fixed structure with four components: (i)~persona framing, (ii)~textual context $C$ from the previous step, (iii)~explanation artifact $e'$, and (iv)~evaluation dimensions $\mathcal{D}$ with the rating scale and constrained response format. The fixed structure supports controlled comparison across datasets, models, XAI explanation methods, and personas.

\vspace{0.2cm}
\noindent\textbf{\protect\circledgreen{5} LLM evaluation.}
% \TODO{this is what you iMHO descrble later, so you can delte it here}
A standardized prompt is provided to the LLM evaluator $\textrm{LLM}_\textrm{judge}(\cdot)$. For visual XAI artifacts, the evaluation input combines the image with its textual context, and thus applies a multimodal LLM.

\vspace{0.2cm}
\noindent\textbf{\protect\circledgreen{6} Statistical analysis.}
The evaluator $\textrm{LLM}_\textrm{judge}(\cdot)$ outputs $(\mathbf{r},\mathbf{j})$, which are stored together with experimental metadata, including dataset, model, XAI explanation method, persona $p$, explanation format, and explanation identifier. The numerical ratings in $\mathbf{r}$ are recorded on a 7-point Likert scale (details below).

We compute descriptive statistics for the ratings in $\mathbf{r}$ by dataset, model, XAI explanation method, persona, explanation format, and evaluation dimension. We assess correlations between evaluation dimensions using Pearson correlations. We compare two-group conditions using two-sided Welch's $t$-tests and multi-group conditions using Welch's ANOVA, followed by Games--Howell post hoc tests where applicable. For synthetic datasets, we use a separate Type-II ANOVA to assess the effects of dataset structure, feature count, and sample size. Statistical significance is assessed at $\alpha = .05$, and effect sizes are reported using $\omega^2$ for Welch's ANOVA and Hedges' $g$ for pairwise comparisons.

% \subsubsection{Illustrative Input--Output Example}

\subsection{Experimental Setup}
\label{sec:exp_setup}
We instantiate the \textsc{XAI-Arena} framework with the following choices for datasets, ML models, XAI explanation methods, evaluation dimensions, stakeholder personas, explanation formats, and instance selection. The evaluation protocol is summarized in Appendix~\ref{app:protocol}.

\subsubsection{Datasets}
We include both (1)~synthetic datasets to allow for controlled experiments and (2)~real-world datasets for realistic evaluation settings. 

$\bullet$\,Synthetic datasets are generated using the \texttt{make\_classification}, \texttt{make\_regression}, \texttt{make\_moons}, and \texttt{make\_circles} functions from \texttt{scikit-learn} \cite{pedregosa2011scikit}. We vary (i)~sample size across \{1{,}000, 3{,}000, 6{,}000, 10{,}000 \}, (ii)~feature count across \{20, 60, 100 \}, and (iii)~data structure (linear vs.\ nonlinear). This design follows established practices for evaluating robustness in XAI research \cite{molnar2020interpretable, alvarez2018robustness, nauta2023anecdotal} and allows us to isolate how dataset characteristics affect how the quality of XAI evaluation is assessed. Further details are reported in Appendices~\ref{app:synth_generation} and~\ref{app:data_validation}. 

$\bullet$\,For real-world benchmarks, we include three tabular datasets covering both classification and regression tasks. The Wisconsin Diagnostic Breast Cancer (WDBC) dataset \cite{wolberg1995breast} contains 569 samples with 30 numerical features describing cell nuclei characteristics derived from digitized biopsy images. It has moderate dimensionality and well-understood feature semantics, and is widely used in interpretability research \cite{lundberg2017unified, ribeiro2016should}.
The Telco Customer Churn dataset contains 7{,}032 samples with 19 original features describing customer demographics, services, and account information. It provides a business-oriented classification task with heterogeneous numerical and categorical features. The California Housing dataset contains 20{,}640 samples with eight numerical features describing housing and demographic characteristics. It provides a real-world regression task with continuous features and a continuous prediction target.

\subsubsection{Preprocessing}
All datasets are split into training, validation, and test sets using a 70/15/15 ratio. Stratified sampling is applied for classification tasks to preserve class distributions across splits. We one-hot encode any categorical features before model training. We standardize numerical features using $z$-score normalization (\texttt{StandardScaler()} in \texttt{scikit-learn}), fitting the scaler on the training data, and applying it to the validation and test sets to avoid data leakage. No additional feature engineering is performed.

% Literature notes: Method - dataset & prepocessing

\subsubsection{Machine Learning Models}

We implement four ML model types: (1)~logistic regression, (2)~random forest, (3)~gradient-boosted trees using XGBoost, and (4)~a neural network implemented as a multi-layer perceptron (MLP). The four models allow us to compare XAI explanation quality across linear, tree-based, and neural network models. We train all models using standard settings from \texttt{scikit-learn} and \texttt{xgboost} \cite{pedregosa2011scikit, chen2016xgboost}, with minor adjustments to improve the stability during training. We generate predictions and XAI explanations for held-out test instances. We do not tune hyperparameters to maintain comparability across models and XAI methods \cite{molnar2020interpretable, carvalho2019machine}. Detailed training settings and validation checks are reported in Appendix~\ref{app:model_training_validation}.

% Literature notes: Method - Machine Learning Models

\subsubsection{XAI Explanation Methods}
\label{sec:xai_methods}

We implement five complementary XAI methods. We focus on three local methods (\textbf{SHAP}, \textbf{LIME}, and \textbf{DiCE}) to generate instance-level explanations, and two global methods (\textbf{PDP} and \textbf{permutation importance}) to summarize model-wide behavior. These methods cover feature attribution, counterfactual explanations, feature importance, and feature--outcome relationships. For classification tasks, SHAP and LIME explain the predicted class.

We export explanation artifacts in standardized textual, tabular, and visual formats to support fair comparison in the LLM evaluation. Feature-based visualizations show only the most salient features. SHAP and LIME use consistent colors for positive and negative feature effects. Details are reported in Appendix~\ref{app:xai_artifact_protocol}.

% Literature notes: Method - Experimental Setup - XAI Explanation Methods

\subsubsection{Evaluation Scope and Instance Selection}
We train each ML model for every applicable dataset configuration, defined by combinations of sample size, feature count, and data structure (linear vs.\ nonlinear). For each dataset--model pair, we randomly sample $n=3$ instances from the held-out test split. Each selected instance is explained using all local XAI methods. We generate each global explanation once per dataset--model configuration. All explanations are evaluated using the same standardized prompt structure.

\subsubsection{Persona Framing}
We evaluate explanations from four stakeholder perspectives $p \in \mathcal{P}$: (1)~ML developer \cite{bhatt2020explainable, liao2021human}, (2)~data scientist \cite{doshi2017towards, hoffman2018metrics}, (3)~manager \cite{miller2019explanation}, and (4)~an end user \cite{gilpin2018explaining, buccinca2020proxy}. Distinguishing these roles reflects the interdisciplinary nature of ML practice, where collaborators differ in professional background and ML knowledge \cite{benk2022turn}. Details are in Table~\ref{tab:personas}. Each persona description is inserted verbatim at the beginning of the prompt and remains fixed across evaluation conditions.

\begin{table}[h]
\centering
\caption{Stakeholder personas, definitions, and prompts used in \textsc{XAI-Arena}.}
\label{tab:personas}
\footnotesize
\renewcommand{\arraystretch}{1.05}
\setlength{\tabcolsep}{4pt}
% \begin{tabular}{p{1.6cm}p{6.6cm}}
\begin{tabular}{
    >{\raggedright\arraybackslash}p{2.8cm}
    >{\raggedright\arraybackslash}p{11cm}
}
\toprule
\textbf{Persona} & \textbf{Definition (and prompt)} \\
\midrule
\parbox[t]{2.6cm}{\textit{ML developer \\
\cite{bhatt2020explainable, liao2021human}}} &
A technical stakeholder focused on debugging models, identifying failure modes, and understanding model internals.\linebreak  
\textbf{Prompt:} \texttt{"You are a machine learning developer. Your goal is to debug the model, see where it goes wrong, and understand how it behaves inside."} \\[2pt]
\midrule
\parbox[t]{2.6cm}{\textit{Data scientist \\
\cite{doshi2017towards, hoffman2018metrics}}} &
A methodological expert interested in assessing feature relevance, statistical reliability, and consistency of XAI explanations.\linebreak  
\textbf{Prompt:} \texttt{"You are a data scientist. You care about which features matter, whether the explanation matches the data and model behavior, and whether it seems statistically reliable."} \\[2pt]
\midrule
\parbox[t]{2.6cm}{\textit{Manager \\
\cite{miller2019explanation}}} &
A decision-maker seeking actionable, decision-supporting explanations that balance clarity, risk, and accountability.\linebreak  
\textbf{Prompt:} \texttt{"You are a manager who has to make decisions based on these results. You want explanations that are easy to understand, show the main risks, and support accountable decisions."} \\[2pt]
\midrule
\parbox[t]{2.6cm}{\textit{End user \\
\cite{gilpin2018explaining, buccinca2020proxy}}} &
A non-technical user who values clear, simple explanations to decide whether to trust and act on model results.\linebreak  
\textbf{Prompt:} \texttt{"You are an end user with no technical background. You just want a clear, simple explanation that helps you decide whether to trust and use the result."} \\
\bottomrule
\end{tabular}
\end{table}

\subsubsection{Evaluation Dimensions}
We evaluate each explanation on the eight dimensions (e.g., \dimension{perceived simplicity}, \dimension{clarity}, \dimension{task adequacy}). Table~\ref{tab:dimensions} lists the definition and exact question used for each dimension.

% Spans one column
\begin{table}[h]
\centering
\caption{Evaluation dimensions, definitions, and questions used in \textsc{XAI-Arena}.}
\label{tab:dimensions}
\footnotesize
\renewcommand{\arraystretch}{1.05}
\setlength{\tabcolsep}{4pt}
% \begin{tabular}{p{1.6cm}p{6.6cm}}
\begin{tabular}{p{2.8cm}>{\raggedright\arraybackslash}p{11.0cm}}
\toprule
\textbf{Dimension} & \textbf{Definition (and prompt)} \\
\midrule
\parbox[t]{2.6cm}{
\dimension{perceived simplicity \\
\cite{hoffman2018metrics, mohseni2021multidisciplinary}}} &
Describes how intuitively an explanation can be understood.\linebreak 
\textbf{Prompt:} \texttt{"How intuitive it is for you to understand the explanation?"} \\[2pt]
\midrule
\parbox[t]{2.6cm}{
\dimension{clarity \\
\cite{hoffman2018metrics, gilpin2018explaining}}} &
Refers to the degree to which the explanation is well structured and unambiguous.\linebreak 
\textbf{Prompt:} \texttt{"How clearly is the explanation presented?"} \\[2pt]
\midrule
\parbox[t]{2.6cm}{
\dimension{task adequacy \\
\cite{hoffman2018metrics, liao2021human}}} &
Indicates whether the explanation provides sufficient information to understand or judge the model’s prediction.\linebreak 
\textbf{Prompt:} \texttt{"Does the explanation give you enough information to understand or judge the ML model’s prediction?"} \\[2pt]
\midrule
\parbox[t]{2.6cm}{
\dimension{trust calibration \\
\cite{buccinca2020proxy, miller2019explanation}}} &
Measures how effectively the explanation helps align user trust with model reliability.\linebreak 
\textbf{Prompt:} \texttt{"How well does the explanation help you adjust your trust in the ML model appropriately (not too much or too little)?"} \\[2pt]
\midrule
\parbox[t]{2.6cm}{
\dimension{actionability \\
\cite{bhatt2020explainable, miller2019explanation, mohseni2021multidisciplinary}}} &
Captures how much the explanation supports meaningful user decisions or actions.\linebreak 
\textbf{Prompt:} \texttt{"If you were using the ML model's prediction to make a decision, how much would the explanation help you take the right next step?"} \\[2pt]
\midrule
\parbox[t]{2.6cm}{
\dimension{transparency \\
\cite{gilpin2018explaining, hoffman2018metrics, liao2021human}}} &
Describes how clearly the explanation reveals the model’s internal logic.\linebreak 
\textbf{Prompt:} \texttt{"How clearly are the ML model's inner workings revealed?"} \\[2pt]
\midrule
\parbox[t]{2.6cm}{
\dimension{faithfulness \\
\cite{doshi2017towards, alvarez2018robustness, vilone2021notions}}} &
Reflects how accurately the explanation represents the model’s reasoning.\linebreak  
\textbf{Prompt:} \texttt{"How closely does the explanation reflect the real reasoning of the ML model?"} \\[2pt]
\midrule
\parbox[t]{2.6cm}{
\dimension{interpretability \\
\cite{doshi2017towards, liao2021human, miller2019explanation}}} &
Encompasses a user’s overall ability to comprehend and reason about the explanation.\linebreak  
\textbf{Prompt:} \texttt{"How understandable is the explanation overall?"} \\
\bottomrule
\end{tabular}
\end{table}

% Literature notes: Method - Experimental Setup - Evaluation dimensions

\subsubsection{LLM Implementation}

We use OpenAI GPT-5.4 (\texttt{gpt-5.4}, accessed April 2026) as the automated evaluator. We assess the cross-LLM robustness in Appendix~\ref{app:cross_llm_robustness}, finding broadly similar patterns. SHAP, LIME, PDP, and permutation importance are evaluated using text-only and text-plus-plot inputs, whereas DiCE remains text-only. We use fixed decoding settings (\texttt{temperature=0}, \texttt{top\_p=1.0}) to reduce sampling variability. The GUIDE-LLM checklist for reporting LLM research \cite{feuerriegel2026reporting} is provided in the supplementary materials.

We implement the pipeline in Python using the packages \texttt{scikit-learn}, \texttt{xgboost}, \texttt{shap}, \texttt{lime}, and \texttt{dice-ml}. PDP and permutation importance are implemented using built-in \texttt{scikit-learn} functions. Each XAI artifact is supplemented with fixed, neutral definitions to describe method-specific terms such as ``contribution'', ``counterfactual'', and ``importance''. The pipeline is version-controlled and executed with fixed random seeds. Evaluation results are exported as structured \texttt{CSV} files containing ratings $\mathbf{r}$, justifications $\mathbf{j}$, and experimental metadata.

% Literature notes: Method - Experimental Setup - Human Evaluation

The final corpus contains 17,252 structured LLM evaluation records, each with ratings and textual justifications for eight dimensions, yielding 138,016 individual dimension ratings. For 19 records, an incorrect output format resulted in one of the eight justifications being missing, but all ratings were complete and thus retained.

%%%%%%%%%%%%%%%%%%%%%%%%%%%%%%%%%%%%%%%%%%%%%%%%%%%%%%%

\section{Results}

Unless otherwise stated, all analyses reported in this section are based on text-only explanations. The effects of explanation format are examined separately in Appendix~\ref{app:format_effects}.

\subsection{Overall LLM-Based Evaluation Patterns}

\researchquestion{1}{NavyBlue}{
Can an LLM distinguish meaningful aspects of XAI explanation quality?
}
We first summarize aggregate ratings and then examine correlations across evaluation dimensions. For simplicity, we interchangeably refer to these perceived explanation quality assessments as ratings below.

\subsubsection{Aggregate Ratings Across Dimensions}

Table~\ref{tab:aggregate_dimension_ratings} reports the mean ($M$)
and standard deviation ($SD$) for each evaluation dimension. The LLM ratings vary across dimensions. \dimension{clarity} receives the highest mean rating, followed by \dimension{perceived simplicity} and \dimension{interpretability}. In contrast, \dimension{transparency} and \dimension{actionability} receive the lowest mean ratings. The explanations are generally rated more highly for being understandable than for revealing model logic or supporting action.

% 4 columns
\begin{table}[h]
\centering
\caption{Aggregate LLM ratings across evaluation dimensions. Ratings are measured on a 1--7 scale.}
\label{tab:aggregate_dimension_ratings}
\footnotesize
\setlength{\tabcolsep}{7pt}
\renewcommand{\arraystretch}{1.10}

\begin{tabular}{@{}lclc@{}}
\toprule
\textbf{Dimension} & \textbf{Mean (SD)}
& \textbf{Dimension} & \textbf{Mean (SD)} \\
\midrule
\dimension{perceived simplicity} & 5.03 (0.95)
& \dimension{actionability}       & 2.89 (0.74) \\

\dimension{clarity}               & 5.70 (0.65)
& \dimension{transparency}        & 3.03 (0.96) \\

\dimension{task adequacy}         & 4.09 (0.64)
& \dimension{faithfulness}        & 4.23 (1.12) \\

\dimension{trust calibration}     & 3.77 (0.66)
& \dimension{interpretability}    & 4.82 (0.79) \\
\bottomrule
\end{tabular}

\vspace{0.3em}
\begin{minipage}{11cm}
\footnotesize
\raggedright
Note: Each dimension is based on 9,988 evaluation records. Values are reported as mean, with standard deviation in parentheses.
\end{minipage}
\end{table}

\subsubsection{Dimension-Level Correlations and Variability}

\begin{figure}[h]
    \centering
    \includegraphics[width=0.70\linewidth]{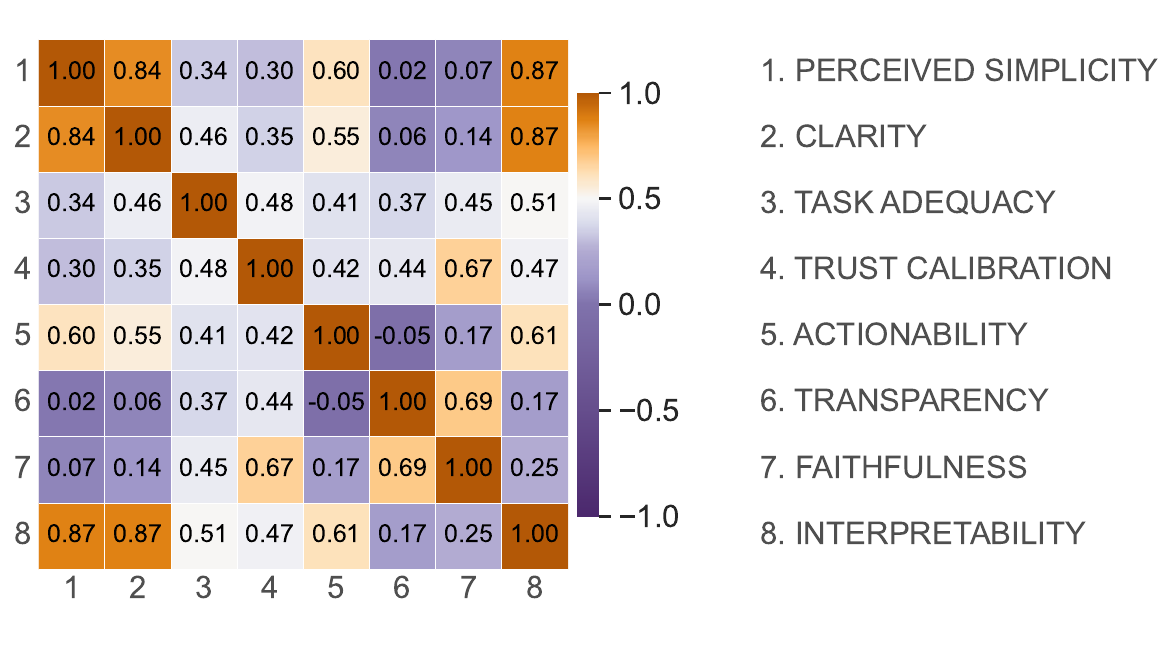}
    \caption{Pearson correlation heatmap across all eight evaluation dimensions.}
    \Description{An eight-by-eight symmetric heatmap showing Pearson correlations between the evaluation dimensions. The strongest correlations occur among perceived simplicity, clarity, and interpretability, ranging from 0.84 to 0.87. Transparency and faithfulness are also strongly correlated at 0.69. Each cell displays its correlation coefficient.}
    \label{fig:dimension_correlation_heatmap}
\end{figure}

Figure~\ref{fig:dimension_correlation_heatmap} shows Pearson correlations among the eight evaluation dimensions. \dimension{clarity}, \dimension{perceived simplicity}, and \dimension{interpretability} are strongly correlated, indicating overlap in comprehensibility. \dimension{faithfulness} and \dimension{transparency} are also closely associated, whereas \dimension{actionability} has weaker correlations with the other dimensions. The dimensions therefore capture related but distinct aspects of explanation quality.

\subsection{Dataset Effects}

\researchquestion{2}{myyellow}{
How sensitive are the ratings to the underlying dataset?
}
We compare how the ratings vary across synthetic vs. real-world datasets as well as across data structure, feature count, and sample size within the synthetic datasets.

\begin{table}[h]
\centering
\caption{LLM ratings for synthetic and real-world datasets across evaluation dimensions. Ratings are measured on a 1--7 scale.}
\label{tab:real_vs_synthetic_dataset_effects}
\footnotesize
\setlength{\tabcolsep}{4pt}
\renewcommand{\arraystretch}{1.10}

\begin{tabular}{@{}lcc@{\hspace{0.5cm}}lcc@{}}
\toprule
\textbf{Dimension}
& \textbf{Synthetic}
& \textbf{Real-world}
& \textbf{Dimension}
& \textbf{Synthetic}
& \textbf{Real-world} \\
\midrule

\dimension{perceived simplicity}
& 5.04 (0.95) & 4.83 (0.87)
& \dimension{actionability}
& 2.90 (0.74) & 2.77 (0.73) \\

\dimension{clarity}
& 5.70 (0.64) & 5.57 (0.66)
& \dimension{transparency}
& 3.03 (0.96) & 3.07 (1.00) \\

\dimension{task adequacy}
& 4.09 (0.62) & 4.11 (0.88)
& \dimension{faithfulness}
& 4.25 (1.11) & 3.84 (1.30) \\

\dimension{trust calibration}
& 3.78 (0.65) & 3.61 (0.92)
& \dimension{interpretability}
& 4.83 (0.78) & 4.69 (0.86) \\

\bottomrule
\end{tabular}

\vspace{0.3em}
\begin{minipage}{11cm}
\footnotesize
\raggedright
Note: Values are reported as mean, with standard deviation in parentheses.
Synthetic datasets contain $n=9{,}504$ ratings per dimension; real-world
datasets contain $n=484$ ratings per dimension.
\end{minipage}
\end{table}

\subsubsection{Real-World vs. Synthetic Dataset Differences}

Table~\ref{tab:real_vs_synthetic_dataset_effects} reports the mean ($M$) and standard deviation ($SD$) of LLM ratings for synthetic and real-world datasets. Ratings are broadly similar across the two dataset origins. Synthetic datasets receive slightly higher mean ratings for \dimension{perceived simplicity}, \dimension{clarity}, \dimension{trust calibration}, \dimension{actionability}, \dimension{faithfulness}, and \dimension{interpretability}, whereas real-world datasets receive slightly higher mean ratings for \dimension{task adequacy} and \dimension{transparency}. The largest mean difference occurs for \dimension{faithfulness}, for which synthetic datasets receive a mean rating of $4.25$ compared with $3.84$ for real-world datasets. Overall, differences by dataset origin appear to be modest, and the rating patterns across dimensions remain largely stable.

%6 columns

\subsubsection{Data Structure, Feature Count, and Sample Size Effects}

\begin{figure}[h]
    \centering
    \includegraphics[width=.92\linewidth]{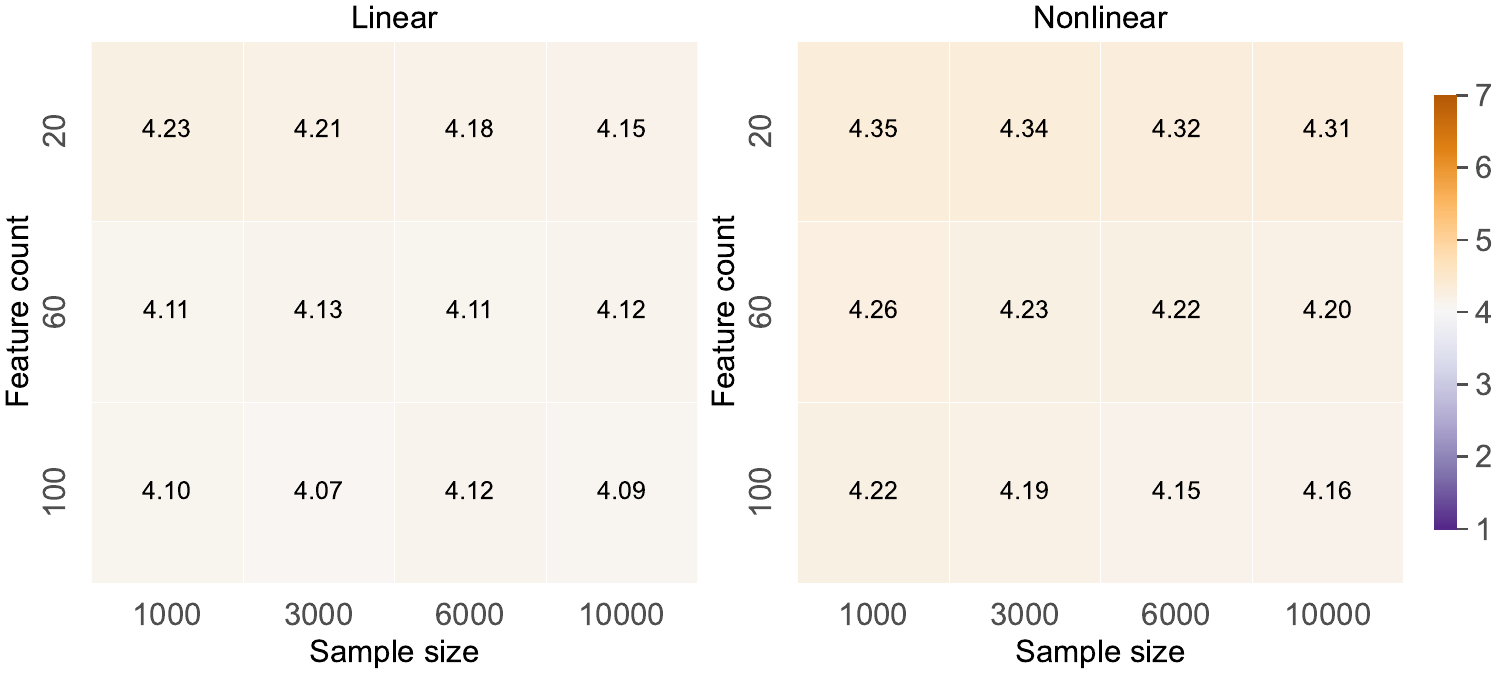}
    \caption{Mean aggregate ratings across synthetic dataset configurations by data structure, feature count, and sample size.}
    \Description{Two side-by-side heatmaps compare mean aggregate ratings for linear and nonlinear synthetic datasets. Rows represent feature counts of 20, 60, and 100, while columns represent sample sizes of 1,000, 3,000, 6,000, and 10,000. Ratings range from 4.07 to 4.35 on the seven-point scale. Nonlinear datasets generally receive slightly higher ratings than linear datasets, configurations with 20 features receive the highest ratings, and differences across sample sizes are small. Each cell displays its mean rating.}
    \label{fig:dataset_structure_heatmap}
\end{figure}

Figure~\ref{fig:dataset_structure_heatmap} shows mean ratings across synthetic dataset configurations by data structure, feature count, and sample size. We use a Type-II ANOVA to quantify the effects of data structure (linear vs.\ nonlinear; $F = 146.03$, $p < .001$), feature count ($F = 74.67$, $p < .001$), and sample size ($F = 4.71$, $p = .003$). Given the large corpus, we focus on the relative magnitude of these patterns rather than statistical significance alone. Data structure and feature count show clearer and more consistent patterns than sample size.

Nonlinear datasets receive slightly higher ratings than linear datasets across most feature count and sample size combinations. Datasets with $20$ features receive the highest ratings, whereas those with $60$ and $100$ features receive lower ratings, in line with expectations. Differences by sample size are small and inconsistent. The absolute rating differences across these dataset characteristics remain modest.

\subsection{Model Effects}

\researchquestion{3}{ForestGreen}{
How sensitive are the ratings to the underlying predictive model?
}

\begin{figure}[h]
    \centering
    \includegraphics[width=0.70\textwidth]
    {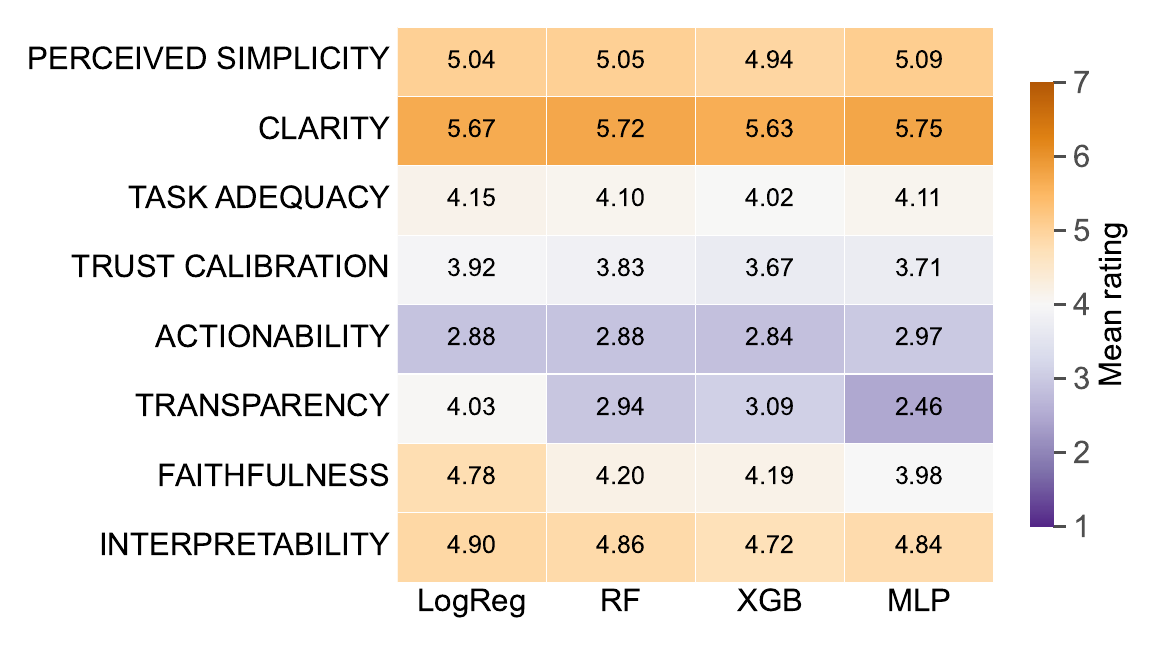}
    \caption{Mean LLM ratings across evaluation dimensions by predictive model.}
    \Description{Heatmap comparing mean ratings across four predictive models and eight explanation-quality dimensions. Differences are most pronounced for transparency and faithfulness. Logistic regression receives the highest ratings for these dimensions, whereas the multilayer perceptron receives lower ratings.}
    \label{fig:model_dimension_heatmap}
\end{figure}

Figure~\ref{fig:model_dimension_heatmap} shows mean ratings across predictive models and evaluation dimensions. Ratings vary only to some degree across predictive models for \dimension{perceived simplicity}, \dimension{clarity}, \dimension{task adequacy}, \dimension{actionability}, and \dimension{interpretability}. Greater variation occurs for \dimension{transparency} and \dimension{faithfulness}. Logistic regression receives the highest ratings on both dimensions, in line with expectations, while the random forest, XGBoost, and particularly MLP receive lower \dimension{transparency} ratings. Welch's ANOVA shows model differences for all eight dimensions ($p < .001$). The largest effect occurs for \dimension{transparency} ($F = 1362.40$, $\omega^2 = .283$), followed by \dimension{faithfulness} ($F = 215.93$, $\omega^2 = .056$). Effects for the remaining dimensions are small, with $\omega^2$ values ranging from $.004$ for \dimension{perceived simplicity} to $.019$ for \dimension{trust calibration}. Thus, ratings show a more pronounced sensitivity for \dimension{transparency} and \dimension{faithfulness}. Detailed statistical results are reported in Appendix~\ref{app:model_effects}.

\subsection{XAI Method Effects}

\researchquestion{4}{BrickRed}{
How does perceived explanation quality vary across XAI methods?
}

\subsubsection{Method-Level Rating Differences}

\begin{figure}
    \centering
    \includegraphics[width=.98\textwidth]{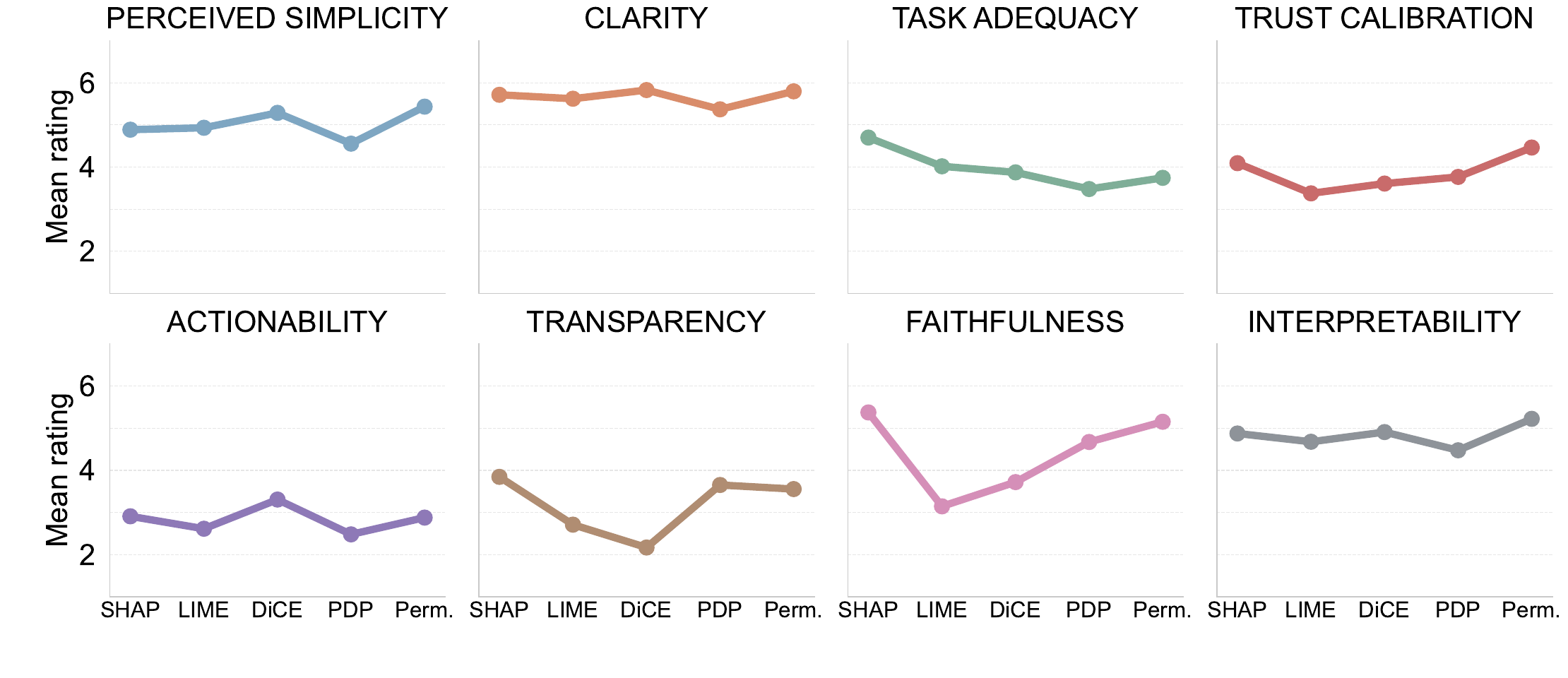}
    \caption{Dimension-specific mean rating patterns across XAI methods.}
    \Description{Eight small-multiple plots show mean ratings on a one-to-seven scale for SHAP, LIME, DiCE, PDP, and permutation importance across perceived simplicity, clarity, task adequacy, trust calibration, actionability, transparency, faithfulness, and interpretability. SHAP receives the highest ratings for task adequacy, transparency, and faithfulness; DiCE for clarity and actionability; and permutation importance for perceived simplicity, trust calibration, and interpretability. Differences are smallest for clarity and most pronounced for transparency and faithfulness.}
    \label{fig:xai_method_pointplot}
\end{figure}

Figure~\ref{fig:xai_method_pointplot} shows mean rating patterns across XAI methods. SHAP receives the highest mean ratings for \dimension{task adequacy}, \dimension{transparency}, and \dimension{faithfulness}. DiCE receives the highest ratings for \dimension{clarity} and \dimension{actionability}; its high \dimension{actionability} rating reflects, in part, the original intention to support what-if decisions \cite{fernandez2022explaining}. Permutation importance receives the highest ratings for \dimension{perceived simplicity}, \dimension{trust calibration}, and \dimension{interpretability}. Differences among methods are relatively small for \dimension{clarity}, but more pronounced for \dimension{faithfulness} and \dimension{transparency}. Welch's ANOVA confirms significant method effects for all eight dimensions ($p < .001$). The largest effect occurs for \dimension{faithfulness} ($\omega^2 = .869$), followed by \dimension{transparency} ($\omega^2 = .814$), \dimension{task adequacy} ($\omega^2 = .653$), and \dimension{trust calibration} ($\omega^2 = .519$). Smaller effects occur for \dimension{actionability} ($\omega^2 = .323$), \dimension{perceived simplicity} ($\omega^2 = .189$), \dimension{interpretability} ($\omega^2 = .135$), and \dimension{clarity} ($\omega^2 = .109$). Thus, perceived explanation quality varies substantially across XAI methods, although the magnitude of this variation differs across evaluation dimensions. Detailed statistical results are reported in Appendix~\ref{app:xai_method_effects}.

\subsubsection{Local vs.\ Global Explanation Patterns}

\begin{table}[h]
\centering
\caption{Mean LLM ratings and Welch's $t$-test results for local
versus global XAI methods across evaluation dimensions.}
\label{tab:local_global_summary}
\small
\setlength{\tabcolsep}{3pt}
\begin{tabular*}{\linewidth}{@{\extracolsep{\fill}}@{}lrrrrc@{}}
\toprule
\textbf{Dimension}
& \multicolumn{2}{c}{\textbf{Local methods}}
& \multicolumn{2}{c}{\textbf{Global methods}}
& \textbf{Test} \\
\cmidrule(lr){2-3}
\cmidrule(lr){4-5}
& \textbf{Mean} & \textbf{SD}
& \textbf{Mean} & \textbf{SD}
& \textbf{$t$ ($p$)} \\
\midrule
\dimension{perceived simplicity} & 5.04 & 0.94 & 4.99 & 1.00 & $1.58$ (.114) \\
\dimension{clarity}              & 5.72 & 0.63 & 5.58 & 0.71 & $7.61$ ($<.001$) \\
\dimension{task adequacy}        & 4.20 & 0.60 & 3.61 & 0.58 & $38.49$ ($<.001$) \\
\dimension{trust calibration}    & 3.69 & 0.64 & 4.11 & 0.66 & $-24.93$ ($<.001$) \\
\dimension{actionability}        & 2.94 & 0.76 & 2.68 & 0.61 & $15.86$ ($<.001$) \\
\dimension{transparency}         & 2.91 & 0.98 & 3.60 & 0.63 & $-38.25$ ($<.001$) \\
\dimension{faithfulness}         & 4.08 & 1.15 & 4.91 & 0.64 & $-42.51$ ($<.001$) \\
\dimension{interpretability}     & 4.82 & 0.75 & 4.85 & 0.95 & $-1.21$ (.227) \\
\bottomrule
\end{tabular*}

\vspace{0.3em}
\begin{flushleft}
\footnotesize
Note: Local methods include SHAP, LIME, and DiCE ($n = 8{,}172$ per dimension); global methods include PDP and permutation importance ($n = 1{,}816$ per dimension). Positive $t$ values indicate higher ratings for local methods; negative values indicate higher ratings for global methods.
\end{flushleft}
\end{table}

Table~\ref{tab:local_global_summary} compares mean ratings for local methods (SHAP, LIME, and DiCE) and global methods (PDP and permutation importance), together with the corresponding Welch's $t$-test results. Local methods receive significantly higher ratings for \dimension{clarity}, \dimension{task adequacy}, and \dimension{actionability}. The largest difference in favor of local methods occurs for \dimension{task adequacy} ($M_{\text{local}} = 4.20$ vs.\ $M_{\text{global}} = 3.61$). Global methods receive significantly higher ratings for \dimension{trust calibration}, \dimension{transparency}, and \dimension{faithfulness}. The largest difference occurs for \dimension{faithfulness} ($M_{\text{global}} = 4.91$ vs.\ $M_{\text{local}} = 4.08$), followed by \dimension{transparency} ($M_{\text{global}} = 3.60$ vs.\ $M_{\text{local}} = 2.91$). No significant differences are found for \dimension{perceived simplicity} ($p = .114$) or \dimension{interpretability} ($p = .227$).

\subsection{Persona Effects}

\researchquestion{5}{TealBlue}{ How does stakeholder perspective shape LLM ratings?}
% \subsubsection{Differences across Stakeholder Roles}

\begin{figure}[h]
    \centering
    \includegraphics[width=0.70\linewidth]
    {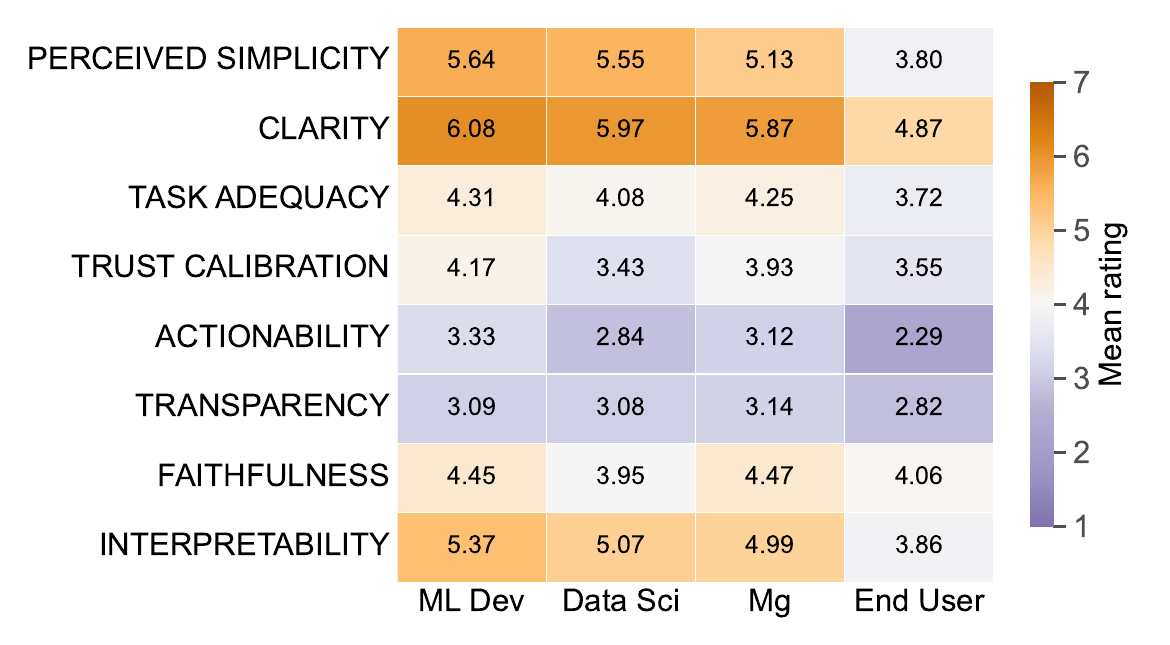}
    \caption{Mean LLM ratings across evaluation dimensions by stakeholder persona.}
    \Description{Heatmap comparing mean ratings across four stakeholder personas and eight explanation-quality dimensions. The end user persona generally receives the lowest ratings, particularly for perceived simplicity, clarity, actionability, and interpretability.}
    \label{fig:persona_dimension_heatmap}
\end{figure}

\begin{figure}
    \centering
    \includegraphics[width=.98\textwidth]{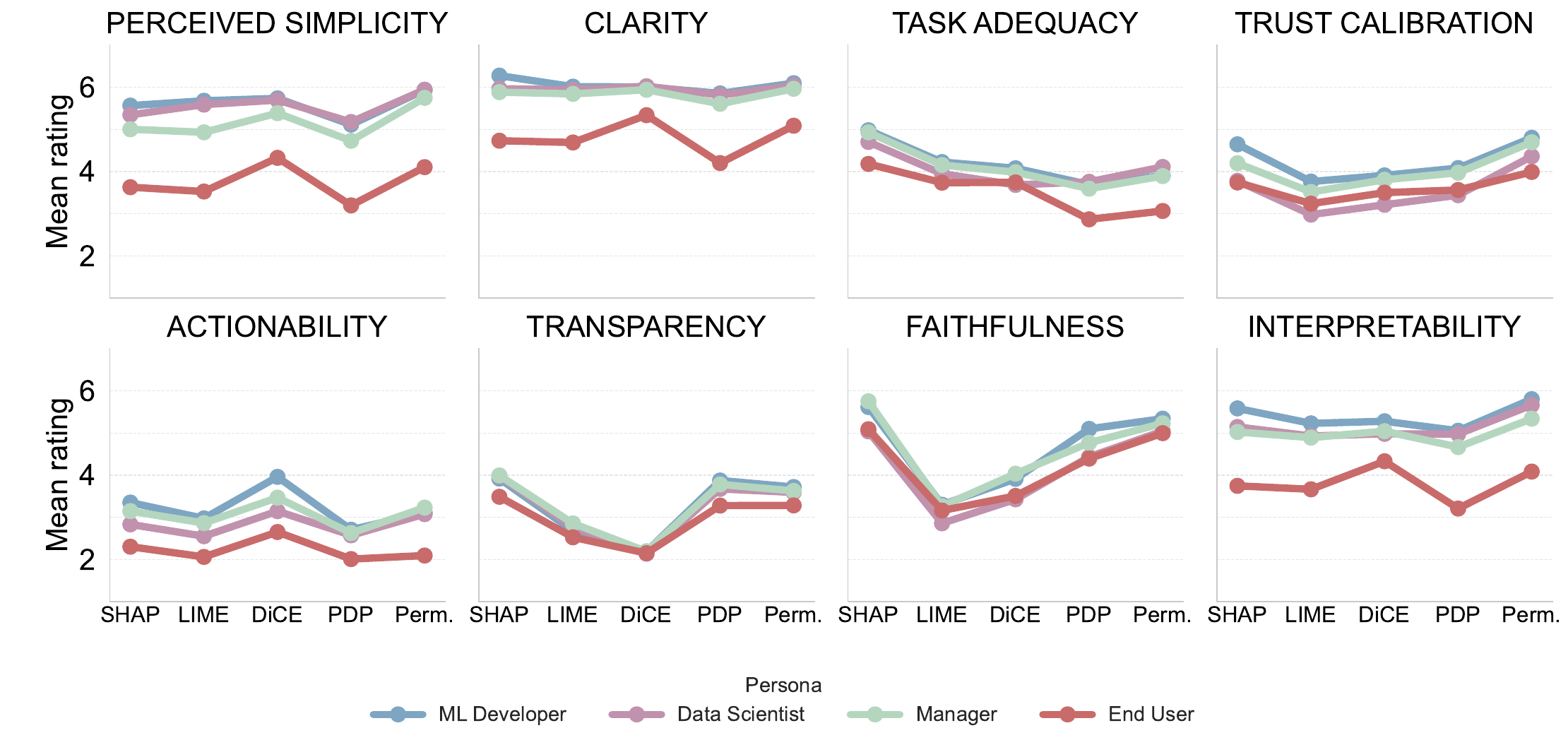}
    \caption{Mean LLM ratings across XAI methods for each stakeholder persona, shown separately by evaluation dimension.}
    \Description{Eight small-multiple plots show mean ratings on a one-to-seven scale for SHAP, LIME, DiCE, PDP, and permutation importance, with separate lines for the ML Developer, Data Scientist, Manager, and End User personas. The End User generally gives the lowest ratings, while the ML Developer and Data Scientist tend to give higher ratings. Persona differences are most pronounced for perceived simplicity, clarity, actionability, and interpretability, and comparatively small for transparency and faithfulness.}
    \label{fig:persona_pointplot_dimensions}
\end{figure}

Figure~\ref{fig:persona_dimension_heatmap} summarizes mean ratings across stakeholder personas and evaluation dimensions, while Figure~\ref{fig:persona_pointplot_dimensions} shows how persona differences vary across XAI methods within each dimension. In line with expectations, the End User persona generally assigns the lowest ratings. Conversely, the ML Developer and Data Scientist personas tend to assign higher ratings. This ordering is broadly consistent across XAI methods, although the magnitude of the persona differences varies by method and dimension. Welch's ANOVA identifies significant persona effects for all eight dimensions ($p < .001$ throughout; Table~\ref{tab:persona_welch_anova}). The largest effects occur for \dimension{perceived simplicity} ($\omega^2 = .711$), \dimension{interpretability} ($\omega^2 = .630$), and \dimension{clarity} ($\omega^2 = .596$). A substantial effect is also observed for \dimension{actionability} ($\omega^2 = .467$), followed by \dimension{trust calibration} ($\omega^2 = .304$) and \dimension{task adequacy} ($\omega^2 = .214$). Persona effects are comparatively small for \dimension{faithfulness} ($\omega^2 = .071$) and \dimension{transparency} ($\omega^2 = .037$). Detailed Games--Howell pairwise comparisons are reported in Appendix~\ref{app:persona_posthoc}.

% 6 columns
\begin{table}[h]
\centering
\caption{Welch's ANOVA results for persona effects across evaluation dimensions.}
\label{tab:persona_welch_anova}
\footnotesize
\setlength{\tabcolsep}{5pt}
\renewcommand{\arraystretch}{1.10}

\begin{tabular}{@{}lrr@{\hspace{0.6cm}}lrr@{}}
\toprule
\textbf{Dimension}
& \textbf{Welch's $F$}
& \textbf{$\omega^2$}
& \textbf{Dimension}
& \textbf{Welch's $F$}
& \textbf{$\omega^2$} \\
\midrule

\dimension{perceived simplicity}
& 4550.29 & .711
& \dimension{actionability}
& 1598.47 & .467 \\

\dimension{clarity}
& 2692.63 & .596
& \dimension{transparency}
& 71.85 & .037 \\

\dimension{task adequacy}
& 504.00 & .214
& \dimension{faithfulness}
& 141.83 & .071 \\

\dimension{trust calibration}
& 807.00 & .304
& \dimension{interpretability}
& 3135.68 & .630 \\

\bottomrule
\end{tabular}

\vspace{0.3em}
\begin{minipage}{11cm}
\footnotesize
\raggedright
\textit{Note.} All tests are significant at $p < .001$.
\end{minipage}
\end{table}

\subsection{External Validation}

\researchquestion{6}{Magenta}{
How well do LLM-based evaluations align with proxy metrics and human annotators?
}

\subsubsection{Alignment with Proxy Metrics}
\label{subsec:proxy_metrics_alignment}

We compare MoRF and stability with the corresponding LLM-rated \dimension{faithfulness} scores for SHAP and LIME. MoRF \cite{samek2016evaluating} sequentially removes features according to their importance ranking and measures the resulting change in model output. The area under the resulting output-drop curve is denoted as MoRF AUC. Stability \cite{alvarez2018robustness} evaluates the consistency of attribution vectors across perturbed instances. We operationalize stability by comparing cosine similarity, Spearman rank stability based on absolute attribution magnitudes, and top-$k$ feature overlap. \dimension{faithfulness} is the primary comparison because the proxy metrics assess technical properties related to explanation faithfulness. \dimension{transparency}, \dimension{trust calibration}, and \dimension{interpretability} are included as reference dimensions.

\begin{figure}[h]
    \centering
    \includegraphics[width=0.70\linewidth]
    {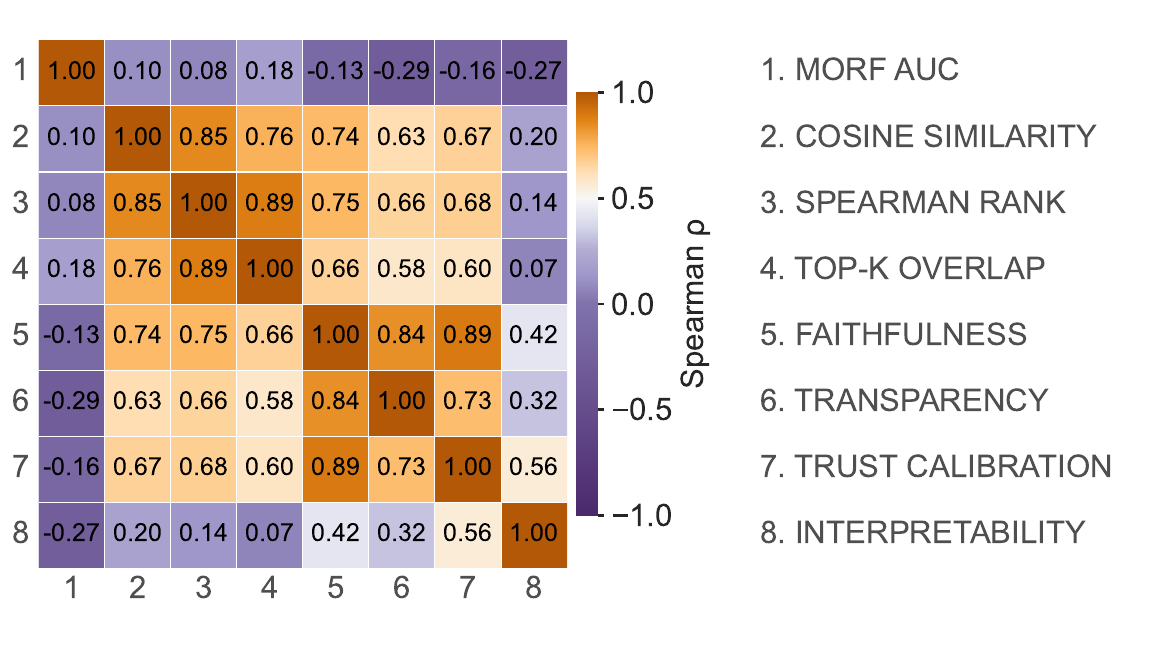}
    \caption{Spearman correlation heatmap between proxy metrics and selected LLM rating dimensions for SHAP and LIME explanations.}
    \Description{An eight-by-eight symmetric heatmap shows Spearman correlations among four proxy metrics—MoRF AUC, cosine similarity, rank stability, and top-k overlap—and four LLM rating dimensions: faithfulness, transparency, trust calibration, and interpretability. The three stability metrics correlate strongly and positively with faithfulness, with coefficients of 0.74, 0.75, and 0.66, whereas MoRF AUC has a weak negative correlation with faithfulness of -0.13. Correlations between the stability metrics and interpretability are comparatively weak. Each cell displays its correlation coefficient, with colors representing values from negative one to positive one.}
    \label{fig:proxy_llm_spearman_heatmap}
\end{figure}

Figure~\ref{fig:proxy_llm_spearman_heatmap} shows Spearman's rank correlation coefficients ($\rho$) between the proxy metrics and the selected LLM rating dimensions. Overall, MoRF AUC shows a very weak negative correlation with \dimension{faithfulness} ($\rho=-.127$, $p=.007$). When analyzed separately, the classification correlation is a very weak negative correlation ($\rho=-.004$, $p=.947$), and the regression correlation is also a very weak negative correlation ($\rho=-.156$, $p=.057$); neither is statistically significant.

By contrast, the stability metrics show strong positive correlations with LLM-rated \dimension{faithfulness}. Rank stability has the strongest association ($\rho=.749$, $p<.001$), followed by cosine similarity ($\rho=.738$, $p<.001$) and top-$k$ overlap ($\rho=.664$, $p<.001$). These results indicate that the stability metrics align more closely with LLM-rated \dimension{faithfulness} than MoRF AUC. However, the stability associations may partly reflect the generally higher stability and \dimension{faithfulness} ratings of SHAP relative to LIME.

\subsubsection{Human--LLM Alignment} 

We recruit $48$ participants aged 18 years or older through Prolific and administer the study in Qualtrics. Participants receive \pounds $6.50$ for an estimated completion time of 20--30 minutes, corresponding to an hourly rate of approximately \pounds $15.60$. Before beginning the survey, participants receive information about the study and provide informed consent, including consent for their anonymized study data to be made publicly available for research purposes. Participation is voluntary, and participants may withdraw at any time before submitting their responses. The survey does not collect names, email addresses, or other direct identifying information. Responses are stored in password-protected project files and analyzed anonymously. The study received ethical clearance from the Ethics Commission at the LMU Munich School of Management (reference: ETH-SOM-075).

Each participant rates six explanation artifacts from a single Telco Customer Churn case. The artifacts cover predictions from MLP and XGBoost models and explanations generated using DiCE and SHAP. Specifically, the survey includes text-only DiCE explanations for MLP and XGBoost, as well as text-only and text+plot SHAP explanations for both models. Participants rate each artifact across all eight evaluation dimensions, yielding 48 ratings per participant. This design supports comparison between human and LLM ratings while keeping the burden for participants manageable. The six evaluation artifacts and the complete survey materials are available in the accompanying supplementary materials.

\begin{figure}[h]
    \centering
    \includegraphics[width=0.70\linewidth]
    {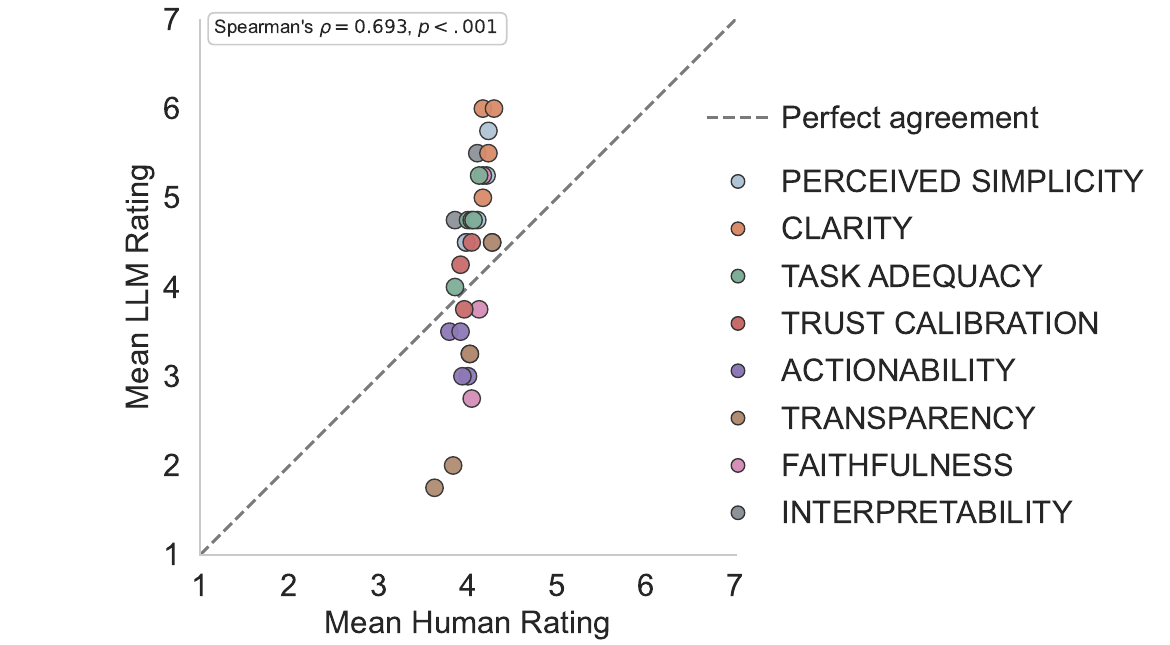}
    \caption{Aggregate alignment between human and LLM evaluations across $32$ text-only artifact--dimension combinations. The dashed $y=x$ line represents perfect agreement.}
    \Description{A scatter plot compares mean human ratings on the horizontal axis with mean LLM ratings on the vertical axis, both ranging from one to seven. The 32 points represent text-only artifact--dimension combinations and are distinguished by evaluation dimension. The points show an overall upward trend, indicating a strong positive association between human and LLM ratings, with Spearman's rho equal to 0.693. The dashed diagonal line represents perfect agreement. Points tend to lie slightly above the diagonal overall, reflecting a small tendency for the LLM to assign higher ratings.}
    \label{fig:human_llm_alignment}
\end{figure}

For the primary role-agnostic alignment analysis, we calculated mean human and LLM ratings for each text-only artifact--dimension combination, aggregating across participants and personas, respectively. The four text-only artifacts and eight evaluation dimensions yielded 32 paired observations. Human and LLM ratings showed a strong positive association (Spearman's $\rho=.693$, $p<.001$). The dimension-specific Spearman correlations showed a very strong positive association for \dimension{perceived simplicity} ($\rho=1.000$), a weak positive association for \dimension{clarity} ($\rho=.389$), a very strong positive association for \dimension{task adequacy} ($\rho=.949$), a weak positive association for \dimension{trust calibration} ($\rho=.200$), a very strong negative association for \dimension{actionability} ($\rho=-.894$), a very strong positive association for \dimension{transparency} ($\rho=1.000$), a very strong positive association for \dimension{faithfulness} ($\rho=.800$), and a strong positive association for \dimension{interpretability} ($\rho=.632$).

As shown in Fig.~\ref{fig:human_llm_alignment}, artifact--dimension combinations receiving higher human ratings generally also received higher LLM ratings. Across all 32 paired observations, the mean absolute error (MAE) was $0.866$, the root mean squared error (RMSE) was $1.003$, and the mean LLM--human difference was $0.255$, indicating a small overall tendency for the LLM to assign higher ratings. Overall, the results demonstrate strong aggregate alignment in relative evaluations.

%%%%%%%%%%%%%%%%%%%%%%%%%%%%%%%%%%%%%%%%%%%%%%%%%%%%%%%

\section{Discussion}

\subsection{Interpretation}

Within the text-only condition, \inlineRQ{1}{NavyBlue} shows that the LLM distinguishes meaningful aspects of explanation quality. Ratings differ across the eight dimensions, while related dimensions follow coherent patterns. In particular, understanding an explanation does not necessarily make it actionable or faithful. This supports prior work that treats XAI quality as a combination of technical and human-centered criteria rather than a single property \cite{doshi2017towards, hoffman2018metrics, liao2021human, nauta2023anecdotal}. \textit{Explanation quality should therefore be evaluated across separate dimensions rather than reduced to one score.}

Turning to dataset and model sensitivity, \inlineRQ{2}{myyellow} shows that ratings follow similar patterns across real-world and synthetic datasets. However, \inlineRQ{3}{ForestGreen} identifies differences between predictive models, which are most pronounced for \dimension{transparency} and, to a lesser extent, \dimension{faithfulness}. The observed pattern is broadly consistent with the expectation that linear models are easier to interpret than more complex models such as neural networks. It suggests that the LLM responds to model properties most clearly when assessing how well an explanation reveals or represents model behavior. This is consistent with earlier findings that explanation quality can depend on the model and data setting in which an XAI method is applied \cite{gilpin2018explaining, arya2019one, covert2021explaining, slack2020fooling}. \textit{Dataset and model choice matter, but they do not affect all aspects of perceived explanation quality equally.}

\inlineRQ{4}{BrickRed} reveals method-specific strengths and weaknesses: no XAI method receives the highest LLM-based rating across all evaluation dimensions. This supports prior work showing that no single method suits every purpose \cite{arya2019one,bhatt2020explainable}. SHAP illustrates this trade-off: despite strong ratings for \dimension{faithfulness} and \dimension{transparency}, it does not consistently lead on more user-centered dimensions. Its use in applied settings \cite{lundberg2017unified,senoner2022using} therefore does not guarantee effective use for every user and task. \textit{Method selection should depend on the explanation goal rather than an overall ranking.}

\inlineRQ{5}{TealBlue} reveals a pronounced stakeholder gap: under the End User persona, the LLM assigns the lowest ratings for \dimension{perceived simplicity}, \dimension{clarity}, \dimension{interpretability}, and \dimension{actionability}. This pattern supports prior research showing that explanation needs to vary with users' expertise, goals, and decision context \cite{miller2019explanation, liao2021human, mohseni2021multidisciplinary, nimmo2024user}. Because XAI can affect human task performance and human--AI collaboration \cite{senoner2022using, senoner2024explainable, schoeffer2024explanations}, explanations rated favorably from a technical perspective may still be unsuitable for their intended users. \textit{XAI evaluation should therefore account for the perspective of the intended stakeholder.}

Finally, \inlineRQ{6}{Magenta} provides converging but qualified evidence for the validity of the LLM-based evaluation. LLM-rated \dimension{faithfulness} shows strong positive associations with the three stability metrics but only a very weak negative association with MoRF AUC. This difference is expected because each proxy captures a specific technical property rather than \dimension{faithfulness} as a whole \cite{alvarez2018robustness, vilone2021notions, nauta2023anecdotal}. The human evaluation provides further support at the aggregate level: explanations rated more highly by human participants also tended to receive higher LLM ratings. However, this strong relative association does not imply exact agreement between human and LLM scores. The dimension-specific correlations varied substantially, suggesting that aggregate alignment may not extend equally across all aspects of explanation quality, although these patterns are exploratory because each coefficient is based on only four artifacts. \textit{Overall, LLM ratings provide a useful aggregate signal of explanation quality, but human judgments, LLM evaluations, and proxy metrics offer complementary rather than interchangeable evidence.}

\subsection{Implications for XAI Benchmarking}

Existing XAI evaluation approaches provide important but incomplete evidence about explanation quality. Human-centered studies capture users' needs and experiences but are costly to scale across many experimental conditions \cite{hoffman2018metrics,liao2021human,mohseni2021multidisciplinary}. Technical proxy metrics enable systematic comparison, but each metric typically captures only a specific property of an explanation \cite{vilone2021notions,nauta2023anecdotal}. Moreover, explanation quality is multidimensional and depends on the intended stakeholder and use context \cite{doshi2017towards,hoffman2018metrics,liao2021human}. \textsc{XAI-Arena} addresses the resulting methodological gap by using LLM-based evaluation to support systematic, multidimensional, and stakeholder-sensitive comparisons of XAI explanations.

Our \textsc{XAI-Arena} framework offers three main methodological strengths. First, it supports scalable comparisons across multiple explanation dimensions rather than relying on a single aggregate score. This makes method-specific strengths and weaknesses visible. Second, it explicitly incorporates stakeholder perspectives, which are often missing from technical proxy benchmarks. Third, it provides a controlled and reproducible protocol through standardized prompts, fixed evaluation dimensions, documented explanation artifacts, and consistent experimental settings. Together, these features make it possible to examine explanation quality systematically while preserving distinctions that conventional aggregate benchmarks may overlook.
 
\textsc{XAI-Arena} is not intended to replace human evaluation. Instead, it provides an additional assessment layer that complements human studies. LLM-based evaluation can support large-scale screening, identify systematic patterns across experimental conditions, and indicate which explanations or stakeholder settings require more focused human investigation. Human evaluation remains necessary to establish whether the LLM-generated outputs correspond to the judgments and experiences of actual stakeholders. We thus envision \textsc{XAI-Arena} as part of a combined evaluation strategy that uses LLM-based assessment to scale XAI benchmarking while retaining targeted human evaluation.

\subsection{Limitations}

Our proposed framework and its empirical evaluation have several limitations. First, the primary evaluation relies on a single LLM, and the ratings may therefore reflect model-specific tendencies. A supplementary cross-LLM check with two additional LLMs examines evaluator dependence and reveals broadly similar degrees of agreement across dimensions (Appendix~\ref{app:cross_llm_robustness}).
Moreover, the selected model is a frontier LLM available at the time of data collection,  providing a strong basis for the primary evaluation. Second, ratings may depend on the selected prompt, persona formulations, and dimension definitions. The prompt was designed following established practices \cite{lin2024write,giray2023prompt,feuerriegel2025using} and standardized across all conditions, which supports internal comparability. Nevertheless, testing additional prompt variants would provide stronger evidence of robustness. Third, LLM outputs may vary across repeated calls, even under fixed decoding settings. Future work should therefore assess the robustness of the findings across multiple evaluation runs. Fourth, the study is limited to tabular data and selected models, XAI methods, and proxy metrics, while the reported comparisons focus on the text-only condition. Nevertheless, the factorial design enables systematic comparison within this scope. Fifth, information leakage from pretraining cannot be excluded, but this risk is reduced because the model evaluates newly generated explanation artifacts rather than reproducing established benchmark answers.

\subsection{Future Work}

Future work could extend \textsc{XAI-Arena} in three directions. First, the framework could be used to assess user needs in high-stakes domains such as medical decision-making, where explanation quality is particularly important for safe use, accountability, and effective communication. Evaluations should represent the distinct perspectives of clinicians, patients, and hospital decision-makers because their knowledge, responsibilities, and explanation needs differ \cite{kuhl2025human}. Second, extending \textsc{XAI-Arena} beyond tabular data to text, time-series, and multimodal models would test whether the evaluation dimensions and stakeholder effects generalize to explanations with different representations and modalities. Third, the framework could support adaptive frameworks, where additional explanations are generated for dimensions that are scored too low (e.g., via the TalkToModel approach \cite{slack2023explaining, bertrand2023interactive}) and thus improve the overall perceived quality. Explanations could also be delivered incrementally to support understanding and memorability~\cite{bo2024incremental}.

%%%%%%%%%%%%%%%%%%%%%%%%%%%%%%%%%%%%%%%%%%%%%%%%%%%%%%%

\section{Conclusion}

Our study shows that LLM-based evaluation can identify systematic differences in the perceived quality of XAI explanations across evaluation dimensions. To the best of our knowledge, \textsc{XAI-Arena} is the first LLM-as-a-judge framework for the multidimensional, human-centered evaluation of XAI explanations. Human ratings show a strong aggregate association with LLM ratings, whereas the relationships with technical proxy metrics depend on the specific property being measured. By combining standardized LLM-based ratings with multiple quality dimensions and stakeholder perspectives, \textsc{XAI-Arena} provides a scalable and reproducible framework for the comparative assessment of XAI explanations.

\section*{AI Use Disclosure}
Generative AI tools are used to assist with the implementation and debugging of research code. All experimental design, code integration, verification, and interpretation of results are performed and controlled by the authors.

\begin{acks}
We thank Foster Provost for the inspiring discussion that let to this idea. Funding by the Deutsche Forschungsgemeinschaft (DFG, German Research Foundation) under the National Research Data Infrastructure – NFDI 27/2, project number 460037581 is acknowledged. The idea emerged from discussion at Dagstuhl Seminar 24342 (Leveraging AI for Management Decision-Making). 
\end{acks}

\bibliographystyle{ACM-Reference-Format}
\bibliography{references}


\clearpage

\appendix

\section{Evaluation Protocol}
\label{app:protocol}

This appendix summarizes the evaluation protocol used to instantiate \textsc{XAI-Arena}. The procedure follows the six-step structure defined in Section~\ref{sec:xai_arena_framework}.

\begin{enumerate}
    \item \textbf{Step 1: Model training.}
    For each dataset configuration, each applicable ML model is trained on the designated training split.

    \item \textbf{Step 2: XAI explanation generation.}
    For each dataset--model pair, $n=3$ test instances $x$ are randomly sampled from the held-out test split. For each selected instance, the model prediction $f(x)$ is computed and local explanation artifacts $e(x,f)$ are generated using the applicable local XAI methods (SHAP, LIME, and DiCE). Global explanation artifacts $e(X_{\mathrm{test}},f)$ are generated once per dataset--model configuration using PDP and permutation importance.
    
    \item \textbf{Step 3: Input construction.}
    For each explanation artifact, the evaluation input combines the artifact with its associated textual context. For local explanations, the context includes the model prediction $f(x)$ and selected feature values of the instance $x$. For global explanations, the context describes the dataset--model setting and the model-level scope of the explanation.
    
    \item \textbf{Step 4: Prompt assembly.}
    The constructed input is embedded into a fixed prompt template under each persona $p \in \mathcal{P}$, including all evaluation dimensions $\mathcal{D}$.
    
    \item \textbf{Step 5: LLM evaluation.}
    Each prompt is submitted to the LLM using fixed decoding settings (\texttt{temperature=0}, \texttt{top\_p=1.0}). The evaluator returns ratings $r$ and textual justifications $j$.
    
    \item \textbf{Step 6: Output logging and statistical analysis.}
    The outputs $(r,j)$ are stored together with experimental metadata (dataset configuration, model, XAI method, persona, instance identifier, and explanation format). Ratings are aggregated to compute summary statistics, statistical tests, and inter-dimension correlations.
\end{enumerate}

All experiments are conducted with fixed random seeds and version-controlled code to support reproducibility.

\section{Example LLM Evaluation Prompt}
\label{app:prompt}

\begin{lstlisting}[basicstyle=\ttfamily\small, breaklines=true]
You are an end user with no technical background. You just want a clear, simple explanation that helps you decide whether to trust and use the result.

Item ID: case_0217

Prediction context:
This task predicts whether a customer is likely to leave the service.
Dataset context: Telco Customer Churn.
The ML model is Logistic Regression.
The model uses 45 input features.
The model predicts class "stay" with probability 0.904.

Case feature values (subset shown):
  PhoneService = Yes
  OnlineBackup = No internet service
  DeviceProtection = No internet service
  PaperlessBilling = No
  TotalCharges = 221.350

Explanation artifact:
DiCE counterfactual explanation for one prediction.
Legend: Counterfactuals are example feature changes that would change the model output.
Number of counterfactuals: 3
Counterfactual changes:
  Counterfactual 1:
    OnlineBackup: No internet service -> Yes
    PaperlessBilling: No -> Yes
    TotalCharges: 221.350 -> 8284.910
  Counterfactual 2:
    PhoneService: Yes -> No
    TotalCharges: 221.350 -> 8461.880
  Counterfactual 3:
    OnlineBackup: No internet service -> Yes
    DeviceProtection: No internet service -> No
    TotalCharges: 221.350 -> 8106.620

Evaluate the explanation across the following dimensions.

For each dimension:
- Provide a score from 1-7 (integer only)
- Provide exactly one concise sentence (max 15 words) for each justification.

Scale definition (for each dimension):
1 = very poor
4 = neutral / neither good nor poor
7 = excellent

Evaluate each dimension independently based only on its definition.

1) Perceived Simplicity - How intuitive it is for you to understand the explanation?
2) Clarity - How clearly is the explanation presented?
3) Task Adequacy - Does the explanation give you enough information to understand or judge the ML model's prediction?
4) Trust Calibration - How well does the explanation help you adjust your trust in the ML model appropriately (not too much or too little)?
5) Actionability - If you were using the ML model's prediction to make a decision, how much would the explanation help you take the right next step?
6) Transparency - How clearly are the ML model's inner workings revealed?
7) Faithfulness - How closely does the explanation reflect the real reasoning of the ML model?
8) Interpretability - How understandable is the explanation overall?

Return only valid JSON in exactly this structure:
{
  "perceived_simplicity": 1,
  "clarity": 1,
  "task_adequacy": 1,
  "trust_calibration": 1,
  "actionability": 1,
  "transparency": 1,
  "faithfulness": 1,
  "interpretability": 1,
  "justifications": {
    "perceived_simplicity": "...",
    "clarity": "...",
    "task_adequacy": "...",
    "trust_calibration": "...",
    "actionability": "...",
    "transparency": "...",
    "faithfulness": "...",
    "interpretability": "..."
  },
  "justification_overall": "..."
}
\end{lstlisting}

\section{Dataset Generation and Validation}
\label{app:data_generation}

\subsection{Synthetic Dataset Generation}
\label{app:synth_generation}

For datasets generated with \texttt{make\_classification} and \texttt{make\_regression}, the number of informative features is fixed at 20, while the remaining features act as independent noise variables, and no redundant features are introduced (\texttt{n\_redundant}=0). Classification datasets use a class separation parameter of \texttt{class\_sep}=1.0, while regression datasets include Gaussian target noise with \texttt{noise}=0.1. For the \texttt{make\_moons} and \texttt{make\_circles} datasets, the two original geometric features define the nonlinear class structure, while additional dimensions are generated as independent Gaussian noise to reach the desired feature dimensionality (\texttt{noise}=0.1).

\subsection{Dataset Validation}
\label{app:data_validation}

All generated dataset splits are validated before downstream experiments. The validation procedure operates at the dataset-directory level and checks both processed split artifacts and associated raw-data artifacts. For each dataset configuration, we first verify the presence of all required processed files, including \texttt{X\_train.npy}, \texttt{X\_val.npy}, \texttt{X\_test.npy}, \texttt{y\_train.npy}, \texttt{y\_val.npy}, \texttt{y\_test.npy}, \texttt{feature\_names.npy}, and \texttt{manifest.json}. We then check structural consistency by confirming that feature and target arrays have matching row counts within each split, that feature dimensionality is consistent across training, validation, and test sets, and that the number of processed feature names matches the observed feature dimension.

In addition, we verify numerical integrity by checking for missing and non-finite values in both feature matrices and target vectors. We also compare the total number of rows across splits against the sample count recorded in the manifest and inspect whether stored metadata such as task type, dataset identifier, and split statistics are internally consistent.

Where available, validation further includes the raw-data artifacts used before preprocessing, including \texttt{X\_train\_raw.csv}, \texttt{X\_val\_raw.csv}, \texttt{X\_test\_raw.csv}, and \texttt{feature\_names\_raw.npy}. For these files, we check split-level row counts, feature dimensionality, consistency of raw feature-name lengths, and agreement between raw and processed split sizes. We also verify the existence of preprocessing-related artifacts and the corresponding raw-data entries in the manifest.

Additional task-specific checks are performed for classification and regression datasets. For classification datasets, we inspect class counts across training, validation, and test splits to identify severe class imbalance or split inconsistencies. For regression datasets, we review summary statistics of the target variable across splits to confirm plausible value ranges and stable distributions. Finally, we check both processed and raw feature-name arrays for duplicate entries.

Together, these checks ensure that all datasets entering subsequent stages of the pipeline are complete, internally consistent, numerically valid, and reproducible. The corresponding validation notebook is available in the code repository.

\section{Model Training, Validation, and Effects}
\label{app:model_training_validation}

\subsection{Model Training}
\label{app:model_training}

All models are trained using the standard implementations provided by \texttt{scikit-learn} and \texttt{xgboost} \cite{pedregosa2011scikit, chen2016xgboost}. To ensure stable and reproducible training across dataset configurations, we apply minor implementation-level adjustments where necessary. For iterative models, we use \texttt{max\_iter=1000} for logistic regression and \texttt{max\_iter=500} for MLP. For logistic regression, we use \texttt{solver=liblinear}. 
For XGBoost, we instantiate the classification and regression variants with \texttt{n\_estimators=200}. We use \texttt{logloss} as the evaluation metric for classification and \texttt{rmse} for regression. No additional hyperparameter tuning is performed in order to preserve comparability across model classes and downstream XAI evaluations.

\subsection{Model Validation}
\label{app:model_validation}

All stored model artifacts are validated after training. For each model bundle, the validation procedure checks file existence, successful loading from disk, and the presence of the expected bundle components, including the trained estimator, feature names, preprocessing entry, and metadata. We further verify whether the estimator appears fitted and whether the corresponding dataset split directory and manifest are available.

Next, we check consistency between the stored model artifact and the associated dataset splits. This includes agreement of the recorded task type and seed with the dataset manifest, agreement between model and split feature dimensionality, and exact alignment between stored feature names and the feature names saved with the dataset splits.

To confirm functional validity, each reloaded model is used to generate predictions on the training, validation, and test splits. In addition, performance metrics are recomputed after reloading and compared against the stored values. For classification models, this includes accuracy; for regression models, this includes \( R^2 \), MAE, and RMSE. Only models that satisfy all structural, consistency, predictive, and metric checks are considered valid for downstream XAI explanation generation and evaluation. The corresponding validation notebook is available in code repository. It regenerates the validation results when the required trained-model artifacts are available.

\subsection{Detailed Statistical Results for Model Effects}
\label{app:model_effects}

Table~\ref{tab:model_effects_welch} reports the Welch ANOVA results for differences among predictive models in the text-only condition. Model effects were statistically significant for all eight evaluation dimensions ($p < .001$). The largest effect occurred for \dimension{transparency} ($\omega^2 = .283$), followed by \dimension{faithfulness} ($\omega^2 = .056$). Effects for the remaining dimensions were small ($\omega^2 = .004$--$.019$).

\begin{table}[h]
\centering
\caption{Welch ANOVA results for predictive-model effects across evaluation
dimensions in the text-only condition.}
\label{tab:model_effects_welch}
\footnotesize
\setlength{\tabcolsep}{4pt}
\begin{tabular}{lrrrr}
\toprule
\textbf{Dimension} & \textbf{Welch's \(F\)} & \(\boldsymbol{df_1}\)
& \(\boldsymbol{df_2}\) & \(\boldsymbol{\omega^2}\) \\
\midrule
\dimension{perceived simplicity} & 12.79   & 3 & 5112.70 & .004 \\
\dimension{clarity}              & 18.03   & 3 & 5131.06 & .005 \\
\dimension{task adequacy}        & 16.50   & 3 & 5111.58 & .005 \\
\dimension{trust calibration}    & 64.66   & 3 & 5113.82 & .019 \\
\dimension{actionability}        & 15.34   & 3 & 5128.63 & .005 \\
\dimension{transparency}         & 1362.40 & 3 & 4783.22 & .283 \\
\dimension{faithfulness}         & 215.93  & 3 & 5079.28 & .056 \\
\dimension{interpretability}     & 21.80   & 3 & 5038.18 & .007 \\
\bottomrule
\end{tabular}
\vspace{0.3em}
\begin{flushleft}
\footnotesize
Note: \(n=9{,}988\) ratings per dimension. All tests were significant at
\(p<.001\). \(\omega^2\) denotes omega squared.
\end{flushleft}
\end{table}

Games--Howell post-hoc tests were used to compare predictive models for \dimension{transparency} and \dimension{faithfulness}. For \dimension{faithfulness}, logistic regression received higher ratings than MLP ($\Delta M = 0.81$, $g = 0.81$, $p < .001$), random forest ($\Delta M = 0.59$, $g = 0.54$, $p < .001$), and XGBoost ($\Delta M = 0.60$, $g = 0.51$, $p < .001$). MLP received lower ratings than random forest ($\Delta M = -0.22$, $g = -0.21$, $p < .001$) and XGBoost ($\Delta M = -0.21$, $g = -0.19$, $p < .001$). Random forest and XGBoost did not differ significantly ($\Delta M = 0.01$, $g = 0.01$, $p = .989$).

For \dimension{transparency}, logistic regression received higher ratings than MLP ($\Delta M = 1.57$, $g = 2.13$, $p < .001$), random forest ($\Delta M = 1.09$, $g = 1.23$, $p < .001$), and XGBoost ($\Delta M = 0.94$, $g = 1.00$, $p < .001$). Among the nonlinear models, MLP received lower ratings than random forest ($\Delta M = -0.48$, $g = -0.69$, $p < .001$) and XGBoost ($\Delta M = -0.63$, $g = -0.84$, $p < .001$), while random forest received slightly lower ratings than XGBoost ($\Delta M = -0.15$, $g = -0.18$, $p < .001$).

\section{XAI Artifact Generation, Validation, and Effects}
\label{app:xai_artifact_protocol}

This appendix provides additional implementation details for the generation and validation of XAI artifacts used in the LLM-based evaluation. All explanation artifacts are generated using fixed random seeds where applicable and are stored together with dataset, model, method, case, and format metadata to ensure traceability.

\subsection{SHAP}

SHAP explanations are generated using model-appropriate explainers. Linear models are explained with \texttt{LinearExplainer}, tree-based models with \texttt{TreeExplainer}, and neural-network models with a generic SHAP explainer. Where no dedicated SHAP explainer is available, approximation-based explanations are used. For classification tasks, SHAP explanations are generated for the predicted class. SHAP artifacts consist of attribution values and visualizations; no additional method-specific interpretive text is added.

\subsection{LIME}

LIME explanations are generated as local feature-weight explanations around the explained instance. Continuous features are discretized, perturbations are sampled around the explained instance, and 1000 perturbed samples are used per explanation. For classification tasks, LIME explanations are generated for the predicted class. LIME visualizations are adapted to use the same red--blue color encoding as SHAP, where red indicates positive and blue indicates negative feature contributions.

\subsection{Permutation Importance}

Permutation importance is computed on the held-out test set using the trained model, test features, and test targets. Feature importance is measured as the decrease in model performance after shuffling each feature. Accuracy is used as the scoring metric for classification tasks, and \( R^2 \) is used for regression tasks. The permutation procedure is repeated 10 times per feature, and the average performance drop is reported. A fixed random seed is used for reproducibility, and parallel computation is enabled where possible.

\subsection{Partial Dependence Plots}

For PDP generation, features are selected using absolute mean permutation importance computed on the test split. PDPs are then generated only for the selected top-\( k \) features. Each PDP curve is computed over a grid of 50 evaluation points. The selected feature-wise PDP curves are aggregated into a single multi-panel figure to provide a consistent global explanation artifact.

\subsection{DiCE}

Counterfactual explanations are generated using the \texttt{dice-ml} library with the random generation method. For classification tasks, counterfactuals are generated for the opposite class, and three counterfactual examples are requested per instance. For regression tasks, desired counterfactual outcomes are defined as shifted prediction intervals. Starting from the model's original prediction, the target is shifted by 10\% of the training-target range. If the original prediction lies in the lower half of the target range, a higher target interval is used; if it lies in the upper half, a lower target interval is used. A narrow interval around the shifted target serves as the desired range for DiCE generation.

\subsection{Visualization Standardization}

For feature-based explanation visualizations, only the top 10 features ranked by the absolute magnitude of the method-specific explanation scores are displayed. This rule is applied to standardize the amount of information shown across explanation artifacts while preserving the most salient features.

\subsection{Artifact Validation}

Before LLM-based evaluation, all XAI artifacts are systematically validated at the artifact level. Across methods, the validation covers artifact existence, successful loading, metadata integrity, consistency with the corresponding dataset splits and model bundles, feature and case alignment, numerical integrity of saved arrays, and plot-file availability. In addition, method-specific checks are performed.

For SHAP and LIME, validation checks whether local explanation artifacts are correctly linked to the corresponding case, dataset, and model, whether saved explanation values and metadata are internally consistent, and whether associated visualizations are present and loadable. For DiCE, validation covers both the explainer directory and the case-level counterfactual artifacts, including the existence of required files, metadata integrity, and consistency with the corresponding dataset and model configuration.

For permutation importance, validation additionally checks consistency of saved importance values, standard deviations, repeated-permutation arrays, scoring metadata, and plot inputs, together with recomputation-based verification against rerun outputs. For PDPs, validation checks directory-level and feature-level metadata, selected-feature consistency, saved response arrays, global and feature-specific artifacts, and consistency between stored plot inputs and the generated figures.

Only artifacts that satisfy the required structural and method-specific checks are considered valid for downstream evaluation. Remaining warnings reflect approximation- or method-related behavior rather than invalid artifact generation. The method-specific validation notebooks are available in the code repository. They regenerate the validation results when the required explanation artifacts are available.

\subsection{Games--Howell Comparisons for XAI Method Effects}
\label{app:xai_method_effects}

Games--Howell post-hoc tests compared methods for \dimension{faithfulness}, \dimension{transparency}, \dimension{actionability}, and \dimension{interpretability}. All pairwise differences reported below were significant at $p < .001$. For \dimension{faithfulness}, SHAP received higher ratings than permutation importance ($\Delta M = 0.22$, $g = 0.32$), PDP ($\Delta M = 0.70$, $g = 0.95$), DiCE ($\Delta M = 1.65$, $g = 2.36$), and LIME ($\Delta M = 2.23$, $g = 3.41$). LIME received the lowest \dimension{faithfulness} rating. For \dimension{transparency}, DiCE received lower ratings than LIME ($\Delta M = -0.54$, $g = -0.95$), PDP ($\Delta M = -1.49$, $g = -3.04$), permutation importance ($\Delta M = -1.39$, $g = -3.28$), and SHAP ($\Delta M = -1.68$, $g = -2.53$). SHAP received the highest \dimension{transparency} rating, whereas DiCE received the lowest. For \dimension{actionability}, DiCE received higher ratings than SHAP ($\Delta M = 0.40$, $g = 0.51$), permutation importance ($\Delta M = 0.43$, $g = 0.49$), LIME ($\Delta M = 0.69$, $g = 0.92$), and PDP ($\Delta M = 0.83$, $g = 0.97$). For \dimension{interpretability}, permutation importance received higher ratings than SHAP ($\Delta M = 0.35$, $g = 0.41$), DiCE ($\Delta M = 0.31$, $g = 0.48$), LIME ($\Delta M = 0.54$, $g = 0.71$), and PDP ($\Delta M = 0.75$, $g = 0.86$). DiCE received a higher rating than LIME ($\Delta M = 0.23$, $g = 0.34$). These comparisons demonstrate distinct method profiles: SHAP received the highest ratings for \dimension{faithfulness} and \dimension{transparency}, DiCE for \dimension{actionability}, and permutation importance for \dimension{interpretability}.

\section{Format Effects}
\label{app:format_effects}

\begin{figure}[h]
    \centering
    \includegraphics[width=.60\linewidth]{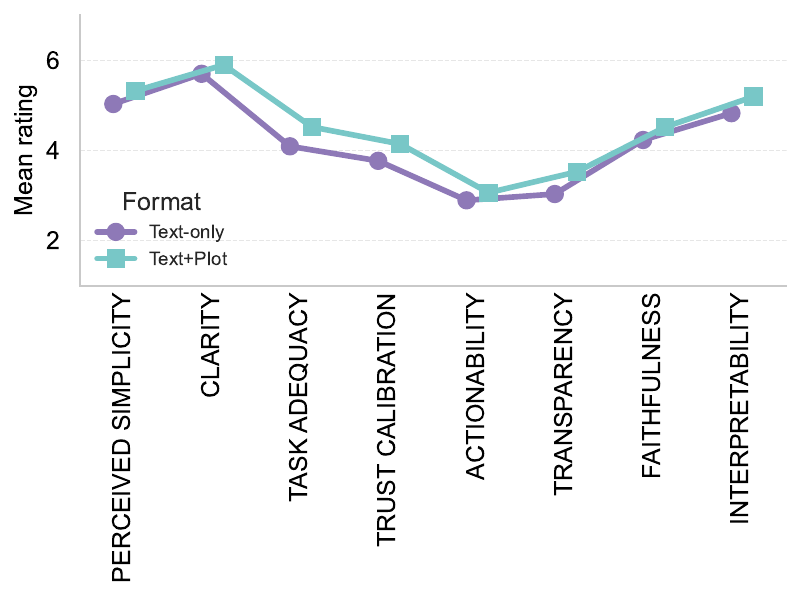}
    \caption{Mean LLM ratings across evaluation dimensions for text-only versus text+plot explanation formats.}
    \Description{A point plot compares mean ratings on a one-to-seven scale for text-only and text-plus-plot explanations across the eight evaluation dimensions. Text-plus-plot explanations receive higher mean ratings than text-only explanations in every dimension, with particularly clear differences for task adequacy, trust calibration, transparency, and interpretability. The two formats nevertheless follow similar rating patterns across dimensions.}
    \label{fig:format_pointplot}
\end{figure}

Figure~\ref{fig:format_pointplot} compares the mean LLM ratings of text-only and text+plot explanations across the eight evaluation dimensions. Text+plot explanations receive consistently higher ratings across all dimensions
($p < .001$ throughout), although the magnitude of the difference varies across dimensions.

To examine whether ratings under the two formats follow similar patterns, we additionally calculated Spearman correlations between text-only and text+plot ratings separately for each XAI method. The correlations were strongly positive for all four methods available in both formats: LIME ($\rho=.957$), SHAP ($\rho=.939$), Permutation Importance ($\rho=.909$), and PDP ($\rho=.885$; all $p<.001$). Thus, although text+plot explanations received consistently higher mean ratings, explanation conditions receiving relatively high ratings in the text-only format also tended to receive relatively high ratings in the text+plot format. DiCE was excluded from this comparison because no text+plot DiCE explanations were available.

\section{Pairwise Comparisons between Stakeholder Personas}
\label{app:persona_posthoc}

Games--Howell tests further examine pairwise persona differences across all eight evaluation dimensions. The largest differences occur for \dimension{perceived simplicity}. ML Developer, Data Scientist, and Manager ratings exceed End User ratings by $\Delta M = 1.84$ ($g = 3.03$), $\Delta M = 1.75$ ($g = 2.80$), and $\Delta M = 1.34$ ($g = 2.12$), respectively. ML Developer ratings also exceed Manager ratings by $\Delta M = 0.51$ ($g = 0.87$), while Data
Scientist ratings exceed Manager ratings by $\Delta M = 0.42$ ($g = 0.69$). The difference between ML Developer and Data Scientist ratings is comparatively small ($\Delta M = 0.09$, $g = 0.16$; all $p < .001$).

For \dimension{clarity}, ML Developer, Data Scientist, and Manager ratings exceed End User ratings by $\Delta M = 1.21$ ($g = 2.45$), $\Delta M = 1.10$ ($g = 2.25$), and $\Delta M = 1.00$ ($g = 2.06$), respectively. Differences among the other three personas are smaller: ML Developer ratings exceed Manager ratings by $\Delta M = 0.21$ ($g = 0.57$) and Data Scientist ratings by $\Delta M = 0.11$ ($g = 0.30$), while Data Scientist ratings exceed Manager ratings by $\Delta M = 0.10$ ($g = 0.28$; all $p < .001$).

For \dimension{task adequacy}, ML Developer ratings exceed End User ratings by $\Delta M = 0.59$ ($g = 1.01$), while Manager and Data Scientist ratings exceed End User ratings by $\Delta M = 0.53$ ($g = 0.89$) and $\Delta M = 0.36$ ($g = 0.59$), respectively. Differences among the ML Developer, Manager, and Data Scientist personas are smaller, ranging from $\Delta M = 0.07$ to $0.23$ ($|g| = 0.12$--$0.39$; all $p < .001$). 

For \dimension{trust calibration}, ML Developer ratings exceed Data Scientist ratings by $\Delta M = 0.74$ ($g = 1.18$) and End User ratings by $\Delta M = 0.62$ ($g = 1.10$). Manager ratings also exceed End User ratings by $\Delta M = 0.39$ ($g = 0.69$), whereas End User ratings slightly exceed Data Scientist ratings by $\Delta M = 0.12$ ($g = 0.19$; all $p < .001$).

For \dimension{actionability}, ML Developer ratings exceed End User ratings by $\Delta M = 1.05$ ($g = 1.66$), while Manager and Data Scientist ratings exceed End User ratings by $\Delta M = 0.83$ ($g = 1.46$) and $\Delta M = 0.55$ ($g = 1.03$), respectively. The remaining pairwise differences are also significant, although their effect sizes are smaller (all $p < .001$).

For \dimension{interpretability}, the largest differences again involve the End User persona. ML Developer ratings exceed End User ratings by $\Delta M = 1.51$ ($g = 2.60$), Data Scientist ratings exceed End User ratings by $\Delta M = 1.21$ ($g = 2.14$), and Manager ratings exceed End User ratings by $\Delta M = 1.12$ ($g = 2.04$). ML Developer ratings also exceed Manager ratings by $\Delta M = 0.39$ ($g = 0.75$) and Data Scientist ratings by $\Delta M = 0.30$ ($g = 0.57$; all $p < .001$).

Differences are smallest for \dimension{transparency} and \dimension{faithfulness}. For \dimension{transparency}, the End User persona assigns lower ratings than the Data Scientist, ML Developer, and Manager personas ($\Delta M = 0.26$--$0.32$, $g = 0.29$--$0.36$; all $p < .001$), while the other three personas do not differ significantly from one another. For \dimension{faithfulness}, Manager and ML Developer ratings exceed Data Scientist and End User ratings. ML Developer and Manager ratings do not differ significantly ($p = .912$), while the small difference between Data Scientist and End User ratings is significant ($\Delta M = -0.11$, $g = -0.10$, $p = .002$).

\section{Cross-LLM Robustness}
\label{app:cross_llm_robustness}

To assess whether the LLM-based ratings depend on the choice of evaluator model, we replayed all 9,988 text-only evaluation prompts using Claude Opus 4.7 and the open-weight Mistral Small 4. We then computed Spearman rank correlations between ratings from the primary GPT-5.4 evaluator and the alternative evaluators for each evaluation dimension. Table~\ref{tab:cross_llm_robustness} reports the pairwise correlations. Overall, the results largely lead to weak positive or strong positive correlations across the dimensions, implying that the different LLMs capture broadly similar patterns. 

\begin{table}[h]
\centering
\caption{Cross-LLM robustness of LLM-based explanation-quality ratings. Values are Spearman rank correlations ($\rho$) across matched text-only evaluation instances ($n=9{,}988$ per dimension).}
\label{tab:cross_llm_robustness}
\small
\begin{tabular}{lccc}
\toprule
Dimension
& GPT--Claude
& GPT--Mistral
& Claude--Mistral \\
\midrule
\textsc{perceived simplicity} & .797*** & .090*** & -.040*** \\
\textsc{clarity}              & .699*** & .288*** & .282*** \\
\textsc{task adequacy}        & .394*** & .570*** & .318*** \\
\textsc{trust calibration}    & .416*** & .268*** & .184*** \\
\textsc{actionability}        & .542*** & .294*** & -.053*** \\
\textsc{transparency}         & .673*** & .597*** & .570*** \\
\textsc{faithfulness}         & .707*** & .569*** & .418*** \\
\textsc{interpretability}     & .673*** & .183*** & -.015 \\
\bottomrule
\multicolumn{4}{l}{\footnotesize $^{*}p<.05$, $^{**}p<.01$, $^{***}p<.001$.}
\end{tabular}
\end{table}

\end{document}